\documentclass[runningheads]{llncs}
\usepackage{amsmath,amssymb} 
\usepackage{graphicx}
\usepackage{times}
\usepackage{epsfig}
\usepackage{color}
\usepackage{algorithm}
\usepackage{algorithmic}
\usepackage{multirow, makecell}
\usepackage[table,xcdraw]{xcolor}
\usepackage[labelsep=period]{caption}
\usepackage{subcaption}
\usepackage{url}

\def\etal{\emph{et al.}}
\def\ETAL{\emph{et al.}}

\usepackage[width=122mm,left=12mm,paperwidth=146mm,height=193mm,top=12mm,paperheight=217mm]{geometry}
\begin{document}
\pagestyle{headings}
\mainmatter
\def\ECCV18SubNumber{681}  

\title{Consistent Multiple Graph Matching with Multi-layer Random Walks Synchronization} 

\titlerunning{Consistent Multiple Graph Matching with Multi-layer Random Walks Synchronization}

\authorrunning{Han-Mu Park \and Kuk-Jin Yoon}

\author{Han-Mu Park$^{1}$ \and Kuk-Jin Yoon$^{2}$}


\institute{Korea Electronics Technology Institute$^{1}$, \\
	Korea Advanced Institute of Science and Technology$^{2}$\\
	\email{hanmu@keti.re.kr$^{1}$,kjyoon@kaist.ac.kr$^{2}$} 
}

\maketitle

\begin{abstract}

We address the correspondence search problem among multiple graphs with complex properties while considering the matching consistency.
We describe each pair of graphs by combining multiple attributes, then jointly match them in a unified framework.
The main contribution of this paper is twofold.
First, we formulate the global correspondence search problem of multi-attributed graphs by utilizing a set of multi-layer structures.
The proposed formulation describes each pair of graphs as a multi-layer structure, and jointly considers whole matching pairs.
Second, we propose a robust multiple graph matching method based on the multi-layer random walks framework.
The proposed framework synchronizes movements of random walkers, and leads them to consistent matching candidates.
In our extensive experiments, the proposed method exhibits robust and accurate performance over the state-of-the-art multiple graph matching algorithms.






\end{abstract}

\section{Introduction}
\label{introduction}
\vspace{-10pt}
 
Graph matching is the problem of finding correspondences between two sets of vertices while preserving complex relational information among them.
Since the graph structure has a strong capacity to represent objects and robustness to severe deformation and outliers, it is frequently adopted to formulate various correspondence problems in the field of computer vision~\cite{conte2004thirty,yan2016short}.
Theoretically, the graph matching problem can be solved by exhaustively searching the entire solution space.
However, this approach is infeasible in practice because the solution space expands exponentially as the size of input data increases. 
For that reason, previous studies have attempted to solve the problem by using various approximation techniques~\cite{cho2010reweighted,cho2014finding,cour2006balanced,gold1996graduated,leordeanu2005spectral,leordeanu2009integer,liu2014gnccp,torresani2013dual,zhang2016pairwise,zhou2012factorized}.
These methods provide satisfactory performance in terms of computational complexity at the expense of some accuracy.
This trade-off between accuracy and complexity is not a serious problem in pairwise graph matching scenarios.
However, this can be crucial in multiple graph matching scenarios because unavoidable errors due to the approximation can accumulate and conflicts can occur between matching pairs.
Therefore, the consistency and accuracy of matching pairs should be considered together in multiple graph matching.

Conventional multiple graph matching methods can be categorized into three types according to the improvement scheme of the matching consistency~\cite{yan2016short}: \textit{iterative}, \textit{one-shot}, and \textit{hybrid} methods.
The \textit{iterative approach} attempts to enhance the consistency by iteratively updating a set of initial solutions.
Yan~\ETAL~\cite{yan2013joint,yan2015consistency} revised the problem formulation as a joint matching problem, and proposed an iterative framework for its optimization.
At the first step, a reference graph is selected, and then each graph pair is replaced with two altered pairs that bypass through the reference graph.
Consequently, the matched vertex pairs automatically satisfy the consistency constraint thanks to the bypass substitution.
Yan~\ETAL~\cite{yan2016multi,yan2014graduated} also proposed a flexible algorithm that gradually improves consistency over iteration.
Park and Yoon~\cite{park2016encouraging} proposed an iterative framework that encourages the soft constraint based on the second-order consistency instead of enforcing the hard constraint of the cycle-consistency.
Since this approach sequentially updates solutions from one graph pair to others, the final results are sensitive to the update sequence, and this often causes the error accumulation.

On the other hand, the \textit{one-shot approach} directly attempts to achieve overall consistency of all pairs of graphs at the same time.
Bonev~\ETAL~\cite{bonev2007constellations} proposed a two-step algorithm to filter out inconsistent matches.
The algorithm finds all possible matching candidates between each graph pair first, and then, iteratively removes the inconsistent matches that violate the cycle-consistency constraint.
Sol{\'e}-Ribalta and Serratosa~\cite{sole2009computation,sole2011models} proposed a more efficient two-step algorithm that uses an N-dimensional probabilistic matrix called a hypercube.
Huang~\ETAL~\cite{huang2012optimization} defined an optimization framework that includes a checking step, in order to maximize the matching consistency and preserve the geometric relationships among neighboring vertices.
These methods filter inconsistent correspondences based on the cycle-consistency constraint to achieve consistency among the multiple graphs.
However, since the consistency constraint is only applied during the post-processing, these methods cannot rectify erroneous initial matching results during the matching process.


To overcome this limitation, the \textit{hybrid approach}~\cite{chen2014near,Maset_2017_ICCV,pachauri2013solving,yan2015matrix,Zhou_2015_ICCV} does not only filter out inconsistent correspondences but also modifies the initial solution based on the information of other matching pairs.
Pachauri~\ETAL~\cite{pachauri2013solving} formulated a global correspondence problem as a problem of finding projections to the universal graph.
Then, the projections to the universal graph are synchronized by using spectral methods.
Maset~\ETAL~\cite{Maset_2017_ICCV} improved the algorithm be robust to the estimation error of the universal graph size.
Chen~\ETAL~\cite{chen2014near} tried to solve the problem by using the semidefinite programming method.
Zhou~\ETAL~\cite{Zhou_2015_ICCV} proposed a method that applies a low-rank relaxation to the set of affinity matrices.
Although the hybrid approach can modify the initial solution to adapt the globally consistent result, but it is still affected by the quality of the input data.
As an answer, Yan~\ETAL~\cite{yan2015matrix} presented a novel formulation based on robust principle component analysis (RPCA)~\cite{candes2011robust} to consider affinity information during the synchronization process.

On the other hand, in an attempt to improve the accuracy of the pairwise graph matching, Park and Yoon~\cite{park2016multi,park2017exploiting,park2017multi} recently addressed the issue of attribute integration which is related to representability of attributes under multi-attributed graph matching settings. 
Many graph matching applications~\cite{cho2010reweighted,cour2006balanced,leordeanu2005spectral,zhou2012factorized,cho2013learning} have used multiple attributes and derived a mixed-type attribute by integration.
This is because the single-type attribute generally does not have enough representability to reflect complex properties of image contents.
However, this integration approach has fundamental problems as stated in \cite{park2016multi,park2017exploiting,park2017multi}.
First, the distinctive information from multiple attributes can be distorted as a consequence of the attribute integration, and 
the distinctive and rich characteristics of one attribute can be weakened by other attributes during the integration process.
Second, the integration recipe cannot be updated or modified adaptively once it is determined. 
This can decrease the flexibility of graph matching because integrated attributes cannot be decomposed again and re-customized for different environments. 
To address this issue, Park and Yoon~\cite{park2016multi,park2017exploiting,park2017multi} proposed a multi-layer structure that jointly represents multiple attributes while preserving its unique characteristics. 
This structure describes each attribute in a separated layer, and connects the layers by using inter-layer edges.
By adjusting the value of inter-layer edges, the relative confidence value among attributes can be redefined during the matching process. 
They also proposed a multi-attributed graph matching algorithm based on the multi-layer structure
, and demonstrated the robust performance over state-of-the-art algorithms.

\begin{figure}[tb]
	\centering
	\includegraphics[width=0.95\linewidth]{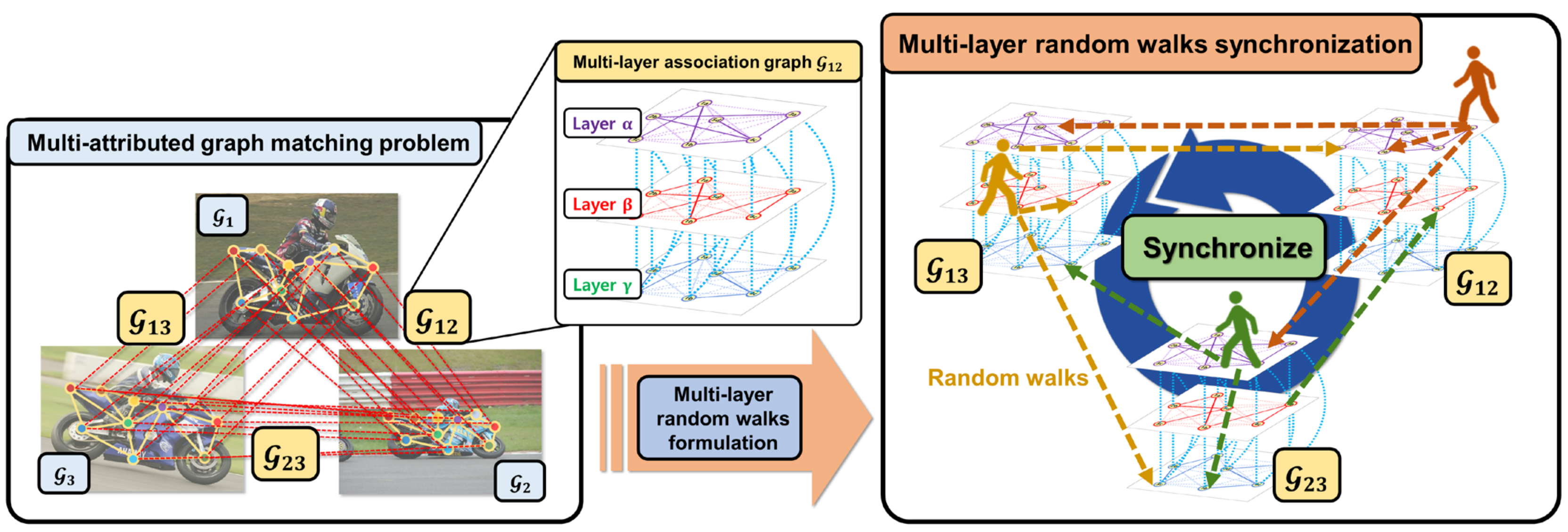} \vspace{-6pt}
	\caption{The proposed multiple graph matching procedure} 
	\label{fig_OverallConcept} \vspace{-15pt}
\end{figure}

In this paper, we propose a robust and consistent multiple graph matching algorithm that considers multiple attributes to deal with general situations in practical applications. 
The proposed method jointly considers affinity information of multiple attributes by using a set of multi-layer association graphs to preserve distinctive information of each attribute and also to find consistent correspondences; moreover, their relations can be adjusted adaptively during the matching process as represented in Fig.~\ref{fig_OverallConcept}.
The main contribution of this paper is twofold.
First, we formulate the global correspondence search problem of multi-attributed graphs by utilizing a set of multi-layer structures.
The proposed formulation describes each pair of graphs as a multi-layer structure, and jointly considers whole matching pairs.
To the best of our knowledge, this is the first attempt to solve the global correspondence search problem of multi-attributed graphs using the multi-layer structure.
Second, we propose a robust multiple graph matching algorithm based on the multi-layer random walks framework.
The proposed method synchronizes the movement of all random walkers, and leads them to consistent matching candidates.
The proposed synchronization approach significantly improves the matching consistency and accuracy over the conventional pairwise multi-layer approach~\cite{park2016multi,park2017exploiting,park2017multi} and achieves robustness against severe deformation by adaptively adjusting contributions of each attribute.

\vspace{-6pt}
\section{Problem Formulation}
\label{problem_formulation}

\vspace{-5pt}
\subsection{Pairwise graph matching problems with multiple attributes}
\label{sec_problem_formulation_pairwise}
\vspace{-3pt}

\begin{figure*}[t!]
	\centering
	\includegraphics[width=\linewidth]{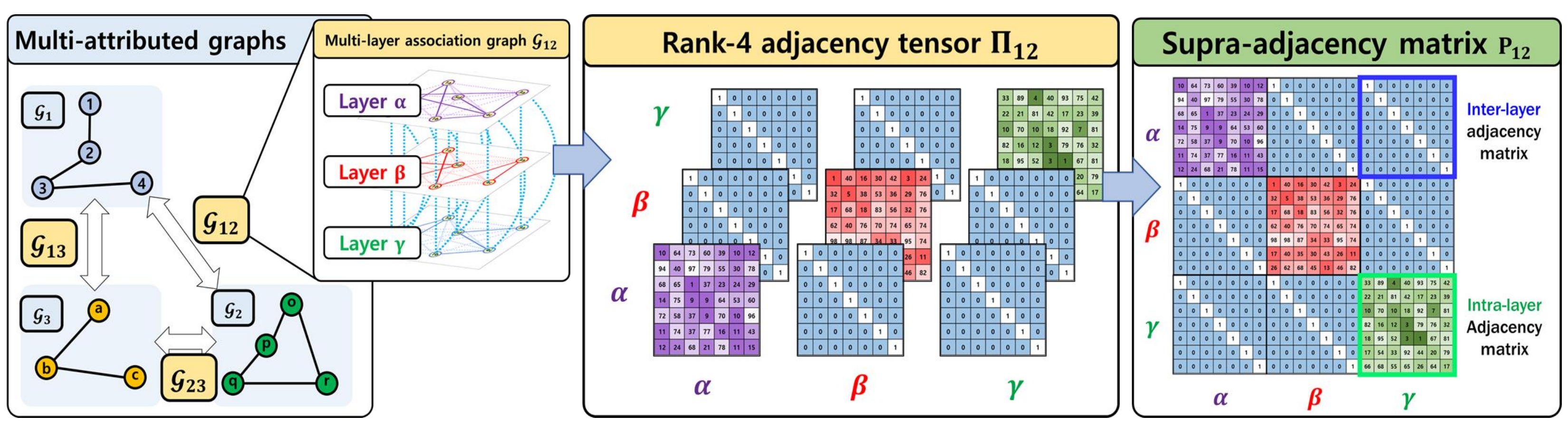}
	\caption{Supra-adjacency matrix $\mathbf{P}_{12}$ is constructed from the multi-layer association graph $\mathcal{G}_{12}$.} 
	\label{fig_ProblemFormulation} 
    \vspace{-10pt}
\end{figure*} 

Suppose two attributed graphs $\mathcal{G}_{1}\left(\mathcal{V}_{1},\mathcal{E}_{1},\mathcal{A}_{1}\right)$ and $\mathcal{G}_{2}\left(\mathcal{V}_{2},\mathcal{E}_{2},\mathcal{A}_{2}\right)$ are given, where $\mathcal{V}$ and $\mathcal{E}$ represent a set of vertices and a set of edges respectively, and $\mathcal{A}$ is a set of attributes which express each vertex and edge.
Then, a correspondence problem between $\mathcal{G}_{1}$ and $\mathcal{G}_{2}$ can be formulated as Lawler's quadratic assignment problem (QAP)~\cite{cho2010reweighted,cour2006balanced,leordeanu2005spectral,zhou2012factorized,park2016multi,park2017exploiting}.
In this formulation, correspondences are represented using an $({N}_{1}\times{N}_{2})$-dimensional binary assignment matrix $\mathbf{X}$, where ${N}_{1}$ and ${N}_{2}$ indicate the numbers of vertices in $\mathcal{G}_{1}$ and $\mathcal{G}_{2}$, respectively.
Each element of the assignment matrix $\left[\mathbf{X}\right]_{i,a}$ represents a correspondence relation of two vertices ${v}_{(1)i}\in\mathcal{V}_{1}$ and ${v}_{(2)a}\in\mathcal{V}_{2}$; if two vertices are matched then the corresponding element is set to 1, otherwise set to 0.
The affinity between correspondences is computed by using the vertex and edge attributes, and is represented by using an $({N}_{1}{N}_{2}\times{N}_{1}{N}_{2})$-dimensional matrix $\mathbf{K}$, which is called an affinity matrix.
For example, a unary affinity of a correspondence pair, ${v}_{(1)i}$ and ${v}_{(2)a}$, is represented in its diagonal element $\left[\mathbf{K}\right]_{ia,ia}$, and a pairwise affinity of two matching candidates $({v}_{(1)i},{v}_{(2)a})$ and $({v}_{(1)j},{v}_{(2)b})$ is represented in a non-diagonal element $\left[\mathbf{K}\right]_{ia,jb}$.
Then, the pairwise graph matching problem is formulated as follows:
\begin{equation}
\begin{split}
\hat { \mathbf{X} } &= \underset { \mathbf{X} } {arg\max } { \  \text{vec}\left(\mathbf{X}\right)^{ \top } \mathbf{K} \ \text{vec}\left(\mathbf{X}\right)  }, \\
s.t. \ \mathbf{X} \in &{ \left\{ 0,1 \right\}  }^{ { N }_{ 1 }\times{ N }_{ 2 } }, \mathbf{X}\mathbf{1}_{{N}_{2}}\le\mathbf{1}_{{N}_{1}}, \mathbf{X}^{\top}\mathbf{1}_{{N}_{1}}\le\mathbf{1}_{{N}_{2}}, 
\end{split}
\label{eq_problem_single_layer_original}
\end{equation}
where $\mathbf{1}_{{N}_{1}}$ indicates an ${N}_{1}$-dimensional all-ones vector, and the inequality constraints denote the one-to-one matching constraints.
As mentioned above, this formulation has the issue arisen from attribute integration when using multiple attributes.

To resolve this issue, the problem can be differently formulated based on a multi-layer association graph as in~\cite{park2016multi,park2017exploiting,park2017multi}. The multi-layer association graph $\mathcal{G}_{k}$ is represented as a set $\left\{\mathcal{V}_{k},\mathcal{E}_{k},\mathcal{A}_{k},\mathcal{L}_{k}\right\}$, where each vertex ${v}^{\alpha}_{(k)ia}\in\mathcal{V}_{k}$ indicates each matching candidate, and each edge ${e}^{\alpha;\beta}_{(k)ia,jb}\in\mathcal{E}_{k}$ represents a pairwise relation between two correspondences.
They regarded $\mathcal{G}_{k}$ as a multi-layer network, and formulated the multi-attributed problem as a centrality problem that finding the most reliable nodes in the network.
The network consists of multiple layers $\mathcal{L}_{k}$ that represent multiple attributes respectively, and the layers are linked to each other as shown in Fig.~\ref{fig_ProblemFormulation}. 
Because of this multi-layer representation, each node ${v}^{\alpha}_{(k)ia}$ has a layer index $\alpha\in\mathcal{L}_{k}$; as a consequence, nodes on different layers, ${v}^{\alpha}_{(k)ia}$ and ${v}^{\beta}_{(k)ia}$, should be distinguished from each other even if they have the same vertex index.
Accordingly, each edge ${e}^{\alpha;\beta}_{(k)ia,jb}$ should have two layer indices.

The edges can be classified into two types: \textit{intra-} and \textit{inter-layer} edges.
The \textit{intra-layer} edge indicates an edge that has the same layer index such as ${e}^{\alpha;\alpha}_{(k)ia,jb}$, and the \textit{inter-layer} edge indicates an edge that has different layer indices such as ${e}^{\alpha;\beta}_{(k)ia,ia}$.
Therefore, a four-dimensional affinity tensor $\mathbf{\Pi}$ is required to describe these two types of relations
, and $\mathbf{\Pi}$ can be flattened to a two-dimensional block affinity matrix $\mathbf{P}$, which is called a supra-adjacency matrix~\cite{park2016multi,park2017exploiting,park2017multi,de2013centrality,gomez2013diffusion,sole2015random}.
Finally, a multi-layer graph matching problem can be formulated by using $\mathbf{P}$ as follows:
\begin{equation}
\begin{split}
\hat{\textbf{X}} &= \underset { \textbf{X} } {arg\max } \left(\textbf{L}^{c}\otimes\text{vec}\left(\textbf{X}\right)\right)^{\top}\textbf{P} \left(\textbf{L}^{c}\otimes\text{vec}\left(\textbf{X}\right)\right) \\
s.t. & \mathbf{X} \in { \left\{ 0,1 \right\}  }^{ { N }_{ 1 }\times{ N }_{ 2 } },  \mathbf{X}\mathbf{1}_{{N}_{2}}\le\mathbf{1}_{{N}_{1}}, \mathbf{X}^{\top}\mathbf{1}_{{N}_{1}}\le\mathbf{1}_{{N}_{2}},
\end{split}
\label{eq_multi_attributed_graph_matching_problem}
\end{equation}
where $\otimes$ indicates the Kronecker product.
$\mathbf{L}^{c}$ is an ${N}_{L}$-dimensional column vector which represents relative confidence values of attributes, and ${N}_{L}$ is the number of layers.

\subsection{Multiple graph matching problems with multiple attributes}
\label{sec_problem_formulation_multiple}

The multiple graph matching problem can be formulated as a summation of individual pairwise matching problems as follows~\cite{yan2013joint,park2016encouraging}:
\begin{equation}
\footnotesize
\begin{array}{c}
\hat{ \mathcal{X} } =\underset { \mathcal{X} } {arg\max }{ \sum\limits _{ l,m=1}^{{N}_{\mathcal{G}}}{ { \text{vec}\left({ \mathbf{X} }_{ lm }\right)  }^{ \top }\left({\mathbf{K}}_{ lm }\right)\text{vec}\left({ \mathbf{X} }_{ lm }\right) }  }, \\
s.t.
\begin{cases}  
\mathcal{X}=\left\{ { \mathbf{X} }_{ 12 },{ \mathbf{X} }_{ 13 },\cdots ,{ \mathbf{X} }_{ ({N}_{\mathcal{G}}-1){N}_{\mathcal{G}} } \right\}, \\ 
{ \mathbf{X} }_{ lm } \in  { \left\{ 0,1 \right\}  }^{ { N }_{ l }\times{ N }_{ m } }, \ l< m \\ 
\mathbf{X}_{ lm }\mathbf{1}_{{N}_{m}}\le\mathbf{1}_{{N}_{l}}, \  {\mathbf{X}_{ lm }}^{\top}\mathbf{1}_{{N}_{l}}\le\mathbf{1}_{{N}_{m}},
\end{cases}
\end{array}
\label{eq_single_layer_MultiIQP}
\end{equation}
where ${N}_{\mathcal{G}}$ is the number of graphs, and $\mathcal{X}$ represents a collection of the individual assignment matrices, $l$ and $m$ are graph indices satisfying $l<m$. 
Since this problem formulation also has the issues arisen form attribute integration, we generalize this formulation to the problem of multi-layer graph matching in order to utilize the rich information of multiple attributes. 
Similar to the single-layer graph matching, the generalized problem can be formulated as a summation of individual multi-layer pairwise matching problems as follows:
\begin{equation}
\footnotesize
\begin{array}{c}
\hat{ \mathcal{X} } =\underset { \mathcal{X} } {arg\max }{ \sum\limits _{ l,m=1}^{{N}_{\mathcal{G}}}{ { \left( \mathbf{L}_{lm}^{c}\otimes\text{vec}\left({ \mathbf{X} }_{ lm }\right) \right)  }^{ \top }\left({\mathbf{P}}_{ lm }\right)\left(\mathbf{L}_{lm}^{c}\otimes\text{vec}\left({ \mathbf{X} }_{ lm }\right)\right) }  }, \\
s.t.
\begin{cases}  
\mathcal{X}=\left\{ { \mathbf{X} }_{ 12 },{ \mathbf{X} }_{ 13 },\cdots ,{ \mathbf{X} }_{ ({N}_{\mathcal{G}}-1){N}_{\mathcal{G}} } \right\}, \\ 
{ \mathbf{X} }_{ lm } \in  { \left\{ 0,1 \right\}  }^{ { N }_{ l }\times{ N }_{ m } }, \  \mathbf{L}^{c}_{lm} \in { \left[ 0,1 \right]  }^{ { N }_{ L } }, \ l< m \\
\mathbf{X}_{ lm }\mathbf{1}_{{N}_{m}}\le\mathbf{1}_{{N}_{l}}, \  {\mathbf{X}_{ lm }}^{\top}\mathbf{1}_{{N}_{l}}\le\mathbf{1}_{{N}_{m}},
\end{cases}
\end{array}
\label{eq_multi_layer_MultiIQP}
\end{equation}
where, ${N}_{L}$ indicate the number of attributes (layers). 
$\mathbf{L}_{lm}^{c}$ is a layer confidence vector of a graph pair $\left(\mathcal{G}_{l},\mathcal{G}_{m}\right)$.

\section{Multi-layer Random Walks Synchronization Method}
\label{proposed_algorithm}

To solve the problem in Eq.~(\ref{eq_multi_layer_MultiIQP}), we propose a multiple graph matching algorithm based on the random walks framework.
The graph matching method based on the framework was firstly proposed by Cho~\etal~\cite{cho2010reweighted}.
In this framework, a matching problem is transformed into the centrality problem of a probabilistic network that consists of matching candidates as nodes.
During the matching process, random walkers randomly traverse the network according to the transition probability computed from the affinity values.
Since the traversal does not consider one-to-one matching constraints, they proposed the reweighting process that leads random walkers to move into nodes which satisfy the constraints.
After converging the traversal, the matching solution is obtained by projecting the stationary distribution of the random walkers into the discrete space.
This method was generalized for the multi-layer association graph to solve the multi-attributed graph matching problem in~\cite{park2016multi,park2017multi}, focused on the possibility of the reweighting process that leads random walkers into desired nodes.
The method, which is called multi-layer random walks matching (MLRWM), utilizes the reweighting process to not only encourage the one-to-one matching constraint but also to control relative confidence values of layers.
We adopt the idea about the reweighting process, because if the random walkers can be led to the nodes that are consistently matched with others, then we can achieve better consistency of the solution. 

\begin{algorithm}[t!]
	\renewcommand{\algorithmicrequire}{\textbf{Input:}}
	\renewcommand{\algorithmicensure}{\textbf{Output:}}
	\caption{Multi-layer Random Walk Synchronization}
	\label{alg_MLSync}
	\begin{algorithmic}[1]   
		\REQUIRE Supra-adjacency matrices $\left\{\mathbf{P}_{lm}\right\}_{l,m=1, l\neq m}^{{N}_{\mathcal{G}}}$, inflation factor $\rho$, reweighting factor $\theta$, reweight synchronizing factor $\omega$, minimum layer confidence value $\tau$, confidence synchronizing factor $\mu$
		\ENSURE Assignment matrices $\hat{\mathcal{X}}$
		
		\STATE Initialize $\forall l,m \ \mathbf{t}_{lm}, \mathbf{u}_{lm}^{sync}, \widetilde{\mathbf{P}}_{lm}$
		\STATE (\textcolor{red}{\textit{Bootstrap process}})
		\STATE Perform multi-layer random walks without synchronization
		\STATE \textbf{repeat}
		\STATE \quad\quad \textbf{for} $l, m = 1$ \textbf{to} ${N}_{\mathcal{G}}$ 
		\STATE \quad\quad\quad\quad (\textcolor{blue}{\textit{Calculate the next distribution}})
		\STATE \quad\quad\quad\quad $\bar{\mathbf{t}}_{lm}^{\top} \leftarrow \mathbf{t}_{lm}^{\top}\widetilde{\mathbf{P}}_{lm}$ 
		\STATE \quad\quad\quad\quad\textbf{for} $\alpha = 1$ \textbf{to} ${N}_{L}$
		\STATE \quad\quad\quad\quad\quad\quad (\textcolor{blue}{\textit{Reweighting random walks for each layer}})
		\STATE \quad\quad\quad\quad\quad\quad $\mathbf{u}_{lm}^{\alpha} \leftarrow \exp(\rho\cdot\bar{\mathbf{t}}_{lm}^{\alpha} / \max(\bar{\mathbf{t}}_{lm}^{\alpha}))$
		\STATE \quad\quad\quad\quad\quad\quad Bistochastic normalize $\mathbf{u}_{lm}^{\alpha}$ using the Sinkhorn method~\cite{sinkhorn1964relationship}
		\STATE \quad\quad\quad\quad\quad\quad (\textcolor{blue}{\textit{Compute Layer-confidence value}})
		\STATE \quad\quad\quad\quad\quad\quad $\left[\mathbf{s}_{lm}\right]_{\alpha} \leftarrow \mathcal{C}_{layer}(\alpha)$
		\STATE \quad\quad\quad\quad \textbf{end}
		\STATE \quad\quad\quad\quad Normalize the layer confidence vector $\mathbf{s}_{lm}$ into $\left[\tau,1\right]$
		\STATE \quad\quad\quad\quad (\textcolor{red}{\textit{Layer confidence synchronization}})
		\STATE \quad\quad\quad\quad $\mathbf{s}_{lm} \leftarrow (1-\mu)\mathbf{s}_{lm} + \mu \ \mathbf{s}^{sync}$
		\STATE \quad\quad\quad\quad (\textcolor{blue}{\textit{Aggregate reweighted distribution}})
		\STATE \quad\quad\quad\quad $\mathbf{u}_{lm}^{temp} \leftarrow \sum_{\alpha} {\left[\mathbf{s}_{lm}\right]_{\alpha} \mathbf{u}_{lm}^{\alpha}}$
		\STATE \quad\quad\quad\quad (\textcolor{red}{\textit{Reweight synchronization}})
		\STATE \quad\quad\quad\quad $\mathbf{u}_{lm}^{temp} \leftarrow (1-\omega)\mathbf{u}_{lm}^{temp} + \omega \ \mathbf{u}_{lm}^{sync}$
		\STATE \quad\quad\quad\quad (\textcolor{blue}{\textit{Diffuse reweighted distribution to whole layer}})
		\STATE \quad\quad\quad\quad $\forall\alpha$ $\mathbf{u}_{lm}^{\alpha} \leftarrow \mathbf{u}_{lm}^{temp}$
		\STATE \quad\quad\quad\quad (\textcolor{blue}{\textit{Update information of reweighted jump}})
		\STATE \quad\quad\quad\quad $\mathbf{t}_{lm} \leftarrow \theta\bar{\mathbf{t}}_{lm} + (1-\theta)\mathbf{u}_{lm}$
		\STATE \quad\quad \textbf{end}
		\STATE \quad\quad (\textcolor{red}{\textit{Layer confidence and reweight synchronization for the next iteration}})
		\STATE \quad\quad Construct synchronized reweight vectors $\mathbf{u}_{lm}^{sync}$
		\STATE \quad\quad Construct synchronized the layer confidence value $\mathbf{s}^{sync}$
		\STATE \textbf{until} $\mathbf{t}$ converges
		\STATE (\textcolor{blue}{\textit{Integrate the assignment vector}})
		\STATE $\text{vec}(\mathbf{X}) \leftarrow \sum_{\alpha} {\mathbf{t}^{\alpha}} $
		\STATE Discretize the assignment matrix $\hat{\mathbf{X}}$
	\end{algorithmic}
\end{algorithm}



Based on this idea, we propose the robust and consistent multiple graph matching algorithm using the multi-layer random walks synchronization. 
The proposed method follows the general random walks framework for graph matching 
as represented in Algorithm~\ref{alg_MLSync}. 
For each iteration, the individual random walker distribution vector $\mathbf{t}_{lm}$, which is obtained by column-wise vectorizing each assignment matrix, is updated according to the predefined transition probability.
The transition probability of each node pair (a correspondence pair) is computed based on the attribute information, and is normalized by using the two-step approach.

Then, each distribution vector is reweighted according to following reweighting criteria that lead random walkers to desired nodes.
First, we give more weight to nodes that satisfy the one-to-one matching constraints by utilizing the inflation and bistochastic normalization steps 
as represented in Line 10-11 of Algorithm~\ref{alg_MLSync}.
The inflation step enlarges the difference between reliable and unreliable matching candidates, and the bistochastic normalization ensures the vector to satisfy the one-to-one matching constraint by using the Sinkhorn method~\cite{sinkhorn1964relationship}.
Second, we assign more weight to reliable attributes based on the layer confidence measure. 
The measure is based on the observation that true correspondences have high affinity with each other than false correspondences. 
This can be defined as the difference in average affinity values between true/false correspondences (see Sec.~\ref{confidence_synchronization} for details).
Last, we give more weights to nodes consistently matched with other pairs for better matching consistency.
This can be implemented by synchronizing the solutions of entire matching pairs. 
However, the consistent nodes cannot be identified during the matching process, because we cannot know the solution.
For that reason, we use the reweighting vectors of each iteration as the temporary solution, then synchronize them by using permutation synchronization methods~\cite{chen2014near,Maset_2017_ICCV,pachauri2013solving} (see Sec.~\ref{reweighting_jump_synchronization} for details).

However, since the random walk distribution is often unstable at the initial stage of the process, the temporary solution could totally different from the optimal solution.
This means the synchronization step may disturb finding consistent nodes, which is the original purpose of this work.
To resolve this issue, we adopt a bootstrap steps at the beginning of the framework. 
The bootstrap steps can stabilize the distribution of random walkers before starting the synchronization, and prevent the entire matching process from falling into the local minima (see Sec.~\ref{bootstrapping} for details).

At the end of the iteration steps, the reweighting vectors $\mathbf{u}_{lm}$ are merged with the current random walker distribution to construct the distribution of the next step.
These steps are iterated until the random walker distribution $\mathbf{t}$ is converged.
Finally, a globally consistent solution can be obtained by binarizing the converged random walker distributions of individual matching pairs using any discretization methods such as the Hungarian method~\cite{munkres1957algorithms}.





\subsection{Layer confidence synchronization}
\label{confidence_synchronization}

The layer confidence value of each matching pair is defined by computing the difference between clusters of true/false correspondences. 
According to the assumption that true correspondences have stronger connections with each other than false correspondences~\cite{leordeanu2005spectral}, the average affinity values of true correspondences should be larger than the value of false correspondences.
Based on this observation, we can define the layer confidence measure using the difference between two mean affinity values of true/false correspondences.
However, since the true/false correspondences cannot be identified during the matching process, we temporarily use the discretized reweighting vector of current iteration as an indicator vector of true correspondences.
Then, the layer confidence values of each matching pair $\mathcal{C}_{layer}\left({\alpha}_{lm}\right)$ are computed as follows:
\begin{equation}
\begin{split}
\mathcal{C}_{layer}\left({\alpha}_{lm}\right) &= \frac{\mathbf{y}_{lm}^{\top}\mathbf{P}_{lm}^{\alpha;\alpha}\mathbf{y}_{lm} - \bar{\mathbf{y}}_{lm}^{\top}\mathbf{P}_{lm}^{\alpha;\alpha}\bar{\mathbf{y}}_{lm}}{2 \cdot \text{Std}(\mathbf{P}_{lm}^{\alpha;\alpha})}\\
s.t. \ \mathbf{y}_{lm} &= \text{Hungarian}(\mathbf{u}_{lm}^{\alpha}),
\end{split}
\end{equation}
where $\mathbf{u}_{lm}$ is the reweighting vector of the matching pair $\left(\mathcal{G}_{l},\mathcal{G}_{m}\right)$, and $\text{Std}(\mathbf{P}_{lm}^{\alpha;\alpha})$ is the standard deviation of $\mathbf{P}_{lm}^{\alpha;\alpha}$ that is adopted to adjust the scale of each attribute.
$\mathbf{y}_{lm}$ is the discretized reweighting distribution that represents the true correspondences, and $\bar{\mathbf{y}}_{lm}$ is the binary complement vector of $\mathbf{y}_{lm}$.
Then, the layer confidence vector $\mathbf{s}_{lm}$ is constructed by normalizing the values based on the measure as follows:
\begin{equation}
\begin{split}
{\left[\mathbf{s}_{lm}\right]}_{\alpha} = \left(1-\tau\right)\frac{\mathcal{C}_{layer}({\alpha}_{lm})}{\mathcal{C}_{max}} + \tau,\\
s.t. \ \mathcal{C}_{max} = \underset{\alpha}{\max} \ {\mathcal{C}_{layer}({\alpha}_{lm})},
\end{split}
\end{equation}
where $\tau$ is a minimum confidence value to ensure the small leverage for each layer even if it is considered as the most unreliable layer.

In this layer confidence synchronization step, the layer confidence vectors of entire pairs are averaged to construct a synchronized confidence vector $\mathbf{s}^{sync}$ as follows:
\begin{equation}
\mathbf{s}^{sync} = \frac{1}{_{{N}_{\mathcal{G}}}{C}_{2}} \sum_{l,m=1, l<m}^{{N}_{\mathcal{G}}}{\mathbf{s}_{lm}}
\label{eq_confidence_synchronization}
\end{equation}
Then, the synchronized vector $\mathbf{s}^{sync}$ is merged with the layer confidence vector of each matching pair $\mathbf{s}_{lm}$.

\subsection{Reweighting jump synchronization}
\label{reweighting_jump_synchronization}

Suppose that all vertices in given graphs belong to a reference graph $\mathcal{G}_{ref}$, and a permutation matrix $\mathbf{U}_{l} \in {\left\{0,1\right\}}^{{N}_{l}\times{N}_{ref}}$ represents projections from a graph $\mathcal{G}_{l}$ to the reference graph $\mathcal{G}_{ref}$.
Then, each assignment matrix can be obtained by multiplying two permutation matrices as follows:
\begin{equation}
\mathbf{X}_{lm} = \mathbf{U}_{l}\mathbf{U}_{m}^{\top}.
\label{eq_permutation_def}
\end{equation}
By using Eq.~(\ref{eq_permutation_def}), the assignment and permutation matrices can be represented as follows:
\begin{equation}
\mathbf{X}=\mathbf{U}\mathbf{U}^{\top},
\label{eq_permutation_combine}
\end{equation}
\begin{equation*}
\begin{array}{c}
\footnotesize 
\mathbf{U}=
\begin{bmatrix}  
\mathbf{U}_{1}  \\  
\mathbf{U}_{2}  \\  
\vdots \\  
\mathbf{U}_{{N}_{\mathcal{G}}}
\end{bmatrix}, \mathbf{X}=
\begin{bmatrix}  
\mathbf{X}_{11} & \mathbf{X}_{12} & \dots & \mathbf{X}_{1{N}_{\mathcal{G}}} \\  
\mathbf{X}_{21} & \mathbf{X}_{22} & \dots & \mathbf{X}_{2{N}_{\mathcal{G}}} \\  
\vdots & \vdots & \ddots & \vdots \\  
\mathbf{X}_{{N}_{\mathcal{G}}1} & \mathbf{X}_{{N}_{\mathcal{G}}2} & \dots & \mathbf{X}_{{N}_{\mathcal{G}}{N}_{\mathcal{G}}}
\end{bmatrix},
\end{array}
\end{equation*}
where the matrix $\mathbf{X}$ has rank ${N}_{ref}$, and is a symmetric positive semidefinite matrix.
Suppose $\widetilde{\mathbf{X}}$ is a block assignment matrix that is estimated from the input data.
Then, the permutation synchronization problem can be formulated as follows:
\begin{equation}
\begin{split}
\hat{ \mathbf{X} } &=\underset { \mathbf{X} } {arg\max }{\left\langle \widetilde{\mathbf{X}},\mathbf{X} \right\rangle } \\
\iff \hat{\mathbf{U}}&=\underset { \mathbf{U} } {arg\max }{\left\langle \widetilde{\mathbf{X}},\mathbf{U}\mathbf{U}^{\top}\right\rangle}\\
&=\underset { \mathbf{U}} {arg\max}{\ tr\left(\mathbf{U}^{\top}\widetilde{\mathbf{X}}\mathbf{U}\right)}, \\
&s.t. \ \mathbf{U}^{\top}\mathbf{U} = \mathbf{I}_{{N}_{\mathcal{G}}},\ \mathbf{X} = \mathbf{U}\mathbf{U}^{\top}.
\end{split}
\label{eq_spectral_formulation}
\end{equation}
Pachauri~\ETAL\cite{pachauri2013solving} proposed a spectral method to solve Eq.~(\ref{eq_spectral_formulation}) by relaxing $\mathbf{U}$ to a continuous matrix.
Since Eq.~(\ref{eq_spectral_formulation}) can be considered as a generalized Rayleigh quotient problem, its solution can be obtained by extracting ${N}_{ref}$ leading eigenvectors of $\widetilde{\mathbf{X}}$.
This spectral method, which is called MatchSync, can solve the permutation synchronization problem in one shot.
However, since the method tries to find actual permutations to the reference graph, the size of the reference graph ${N}_{ref}$ should be correctly estimated.
To resolve this limitation, Maset~\ETAL\cite{Maset_2017_ICCV} proposed the modified algorithm that directly synchronizes pairwise assignment matrices $\mathbf{X}$.
The method, which is called MatchEIG, only needs to know the minimum size of the reference graph to ensure proper performance while MatchSync needs to know the exact size.
At the first step of MatchEIG, ${N}_{ref}$ leading eigenvectors of $\widetilde{\mathbf{X}}$ are integrated to construct an approximated permutation $\widetilde{\mathbf{U}}$, and corresponding eigenvalues are collected to construct a diagonal matrix $\mathbf{D}$.
Then, the synchronized block assignment matrix $\hat{\mathbf{X}}$ can be obtained as follows:
\begin{equation}
\hat{\mathbf{X}} = \widetilde{\mathbf{U}}\mathbf{D}\widetilde{\mathbf{U}}^{\top}.
\end{equation}
Since $\hat{\mathbf{X}}$ is a continuous matrix, any discretization method, such as Hungarian method~\cite{munkres1957algorithms}, should be adopted to obtain the binary solution.

The proposed algorithm adopts one of the permutation synchronization methods, MatchEIG~\cite{Maset_2017_ICCV}, to synchronize the reweighting vectors of the current iteration.
First, a block assignment matrix $\widetilde{\mathbf{X}}$ is constructed by integrating the reweighting vectors of whole pairs, and then is fed to MatchEIG as the input. 
Finally, the synchronized reweighting vectors can be obtained by separating $\hat{\mathbf{X}}$, and combined with the reweighting vectors of each matching pair.
Note that any permutation synchronization methods can be used for this step.
Although we select MatchEIG~\cite{Maset_2017_ICCV} in this paper because of its good balance between accuracy and efficiency, any other methods such as MatchSync~\cite{pachauri2013solving} or MatchLift~\cite{chen2014near} can be adopted.


\subsection{Bootstrapping}
\label{bootstrapping}

Since the distribution of random walkers is often unstable at the initial stage of the process, the random walkers can be biased towards unreliable matching candidates if the synchronization is performed based on the unstable information.
For that reason, we incorporate a bootstrap process at the initial stage (Line 3 of Algorithm~\ref{alg_MLSync}), which does not synchronize the reweighting vectors and layer confidence values.
By waiting the random walker distribution until being stabilized through the bootstrap process, it can prevent the entire matching process from falling into the local minima.


\section{Experimental Results}
\label{experimental_results}


To evaluate the proposed algorithm, we performed two experiments using the synthetic and WILLOW datasets~\cite{cho2013learning}.\footnote{http://www.di.ens.fr/willow/research/graphlearning/}
We compare our algorithm with the well-known pairwise graph matching algorithms such as reweighted random walks matching (RRWM)~\cite{cho2010reweighted}, and multi-layer random walk matching (MLRWM)~\cite{park2016multi}, and multiple graph matching algorithms such as MatchOpt (MOpt)~\cite{yan2013joint,yan2015consistency}, MatchSync (MSync)~\cite{pachauri2013solving}, Composition based affinity optimization (CAO)~\cite{yan2016multi,yan2014graduated}, MatchLift (MLift)~\cite{chen2014near}, MatchALS (MALS)~\cite{Zhou_2015_ICCV}, and MatchEIG (MEIG)~\cite{Maset_2017_ICCV}.

In all experiments, we fix the reweighting factor $\theta$ as 0.2, the minimum layer confidence value $\tau$ as 0.1, the reweighting jump synchronizing factor $\omega$ as 0.8, and the layer confidence synchronizing factor $\mu$ as 0.5.
The inflation factor $\rho$ is set to 100 for all datasets, and each experiment is iterated 50 times.
The parameters of compared algorithms are selected as provided in the original papers, and RRWM is adopted as a pairwise solver to generate the input matching results for multiple graph matching algorithms that require initial solutions.
Our evaluation framework is based on the open MATLAB programs of \cite{cho2010reweighted} and \cite{park2016multi}; and the other matching algorithms are used from the authors' open source codes.

\subsection{Performance evaluation on synthetic dataset}
\label{synthetic_experiments}


\begin{table*}[t]
	\centering
	\caption{Parameter setting for the synthetic graph matching experiments}
	\label{tab_param_synthetic}
	{
    	\small
		\begin{tabular}{l|c|c}
			\noalign{\hrule height 0.5pt} \hline			
			Experiments      & Varied parameter                    & Fixed parameters                                    \\ \hline		\hline
			Deformation      &    $\epsilon=0$ -- $0.3$     &  ${N}_{\mathcal{G}}=10$, ${N}_{att}=5,10$, ${N}_{in}=10$, ${N}_{out}=2$, ${\sigma}^{2}=0.3$   \\
			Outlier          &    ${N}_{out}=0$ -- $10$     &  ${N}_{\mathcal{G}}=10$, ${N}_{att}=5,10$, ${N}_{in}=10$, $\epsilon=0.1$, ${\sigma}^{2}=0.3$  \\
			Graph set size	 	 &    ${N}_{\mathcal{G}}=4$ -- $20$    &  ${N}_{att}=5,10$, ${N}_{in}=10$, ${N}_{out}=2$, $\epsilon=0.1$, ${\sigma}^{2}=0.3$     \\ 
			\noalign{\hrule height 0.5pt} \hline	
		\end{tabular}
	}
    \vspace{-8pt}
\end{table*}

\begin{figure*}[t!]
  \begin{subfigure}[t]{\textwidth}
    \includegraphics[width=0.42\linewidth]{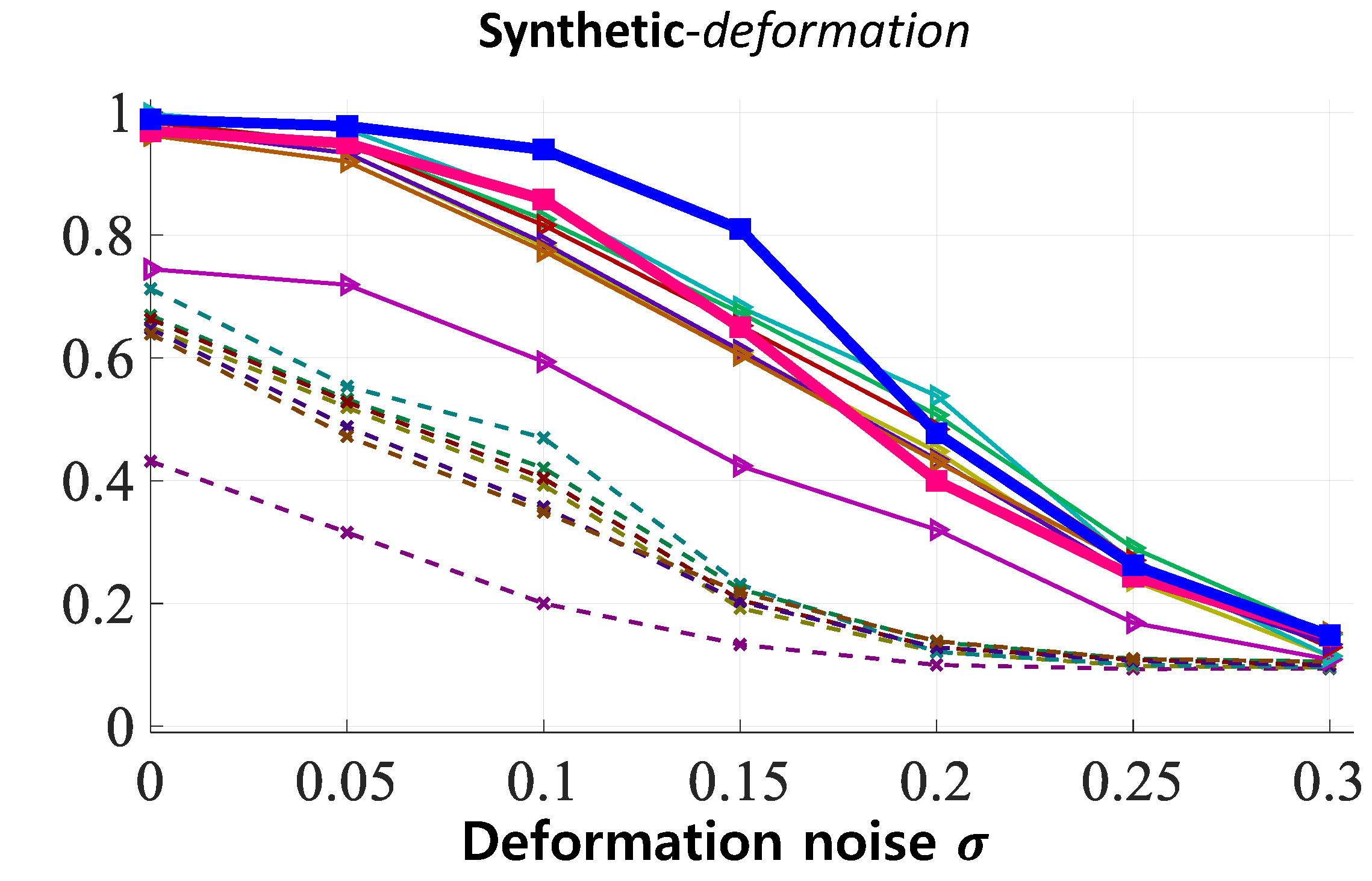}
    \includegraphics[width=0.42\linewidth]{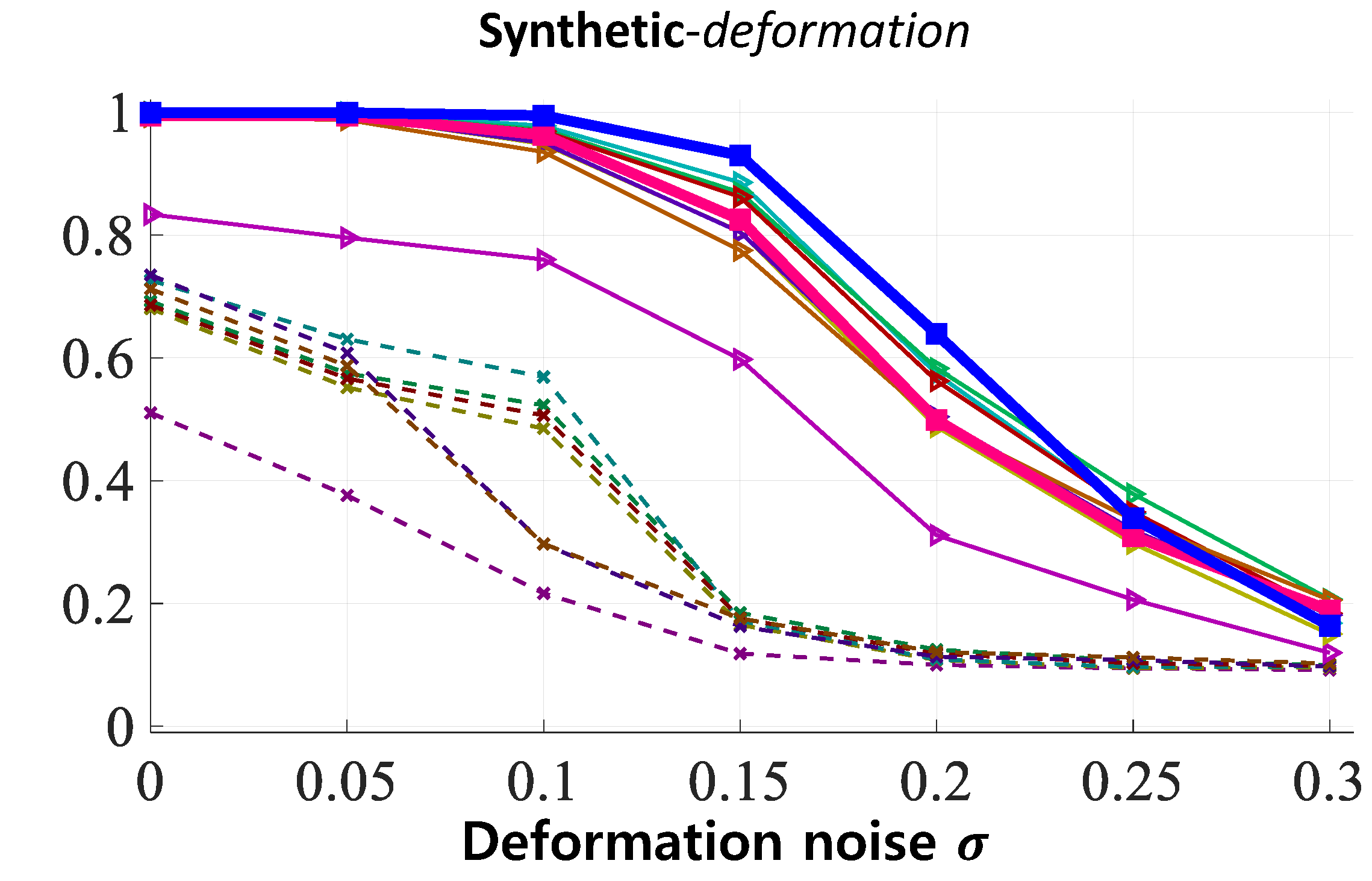}
    \includegraphics[width=0.15\linewidth]{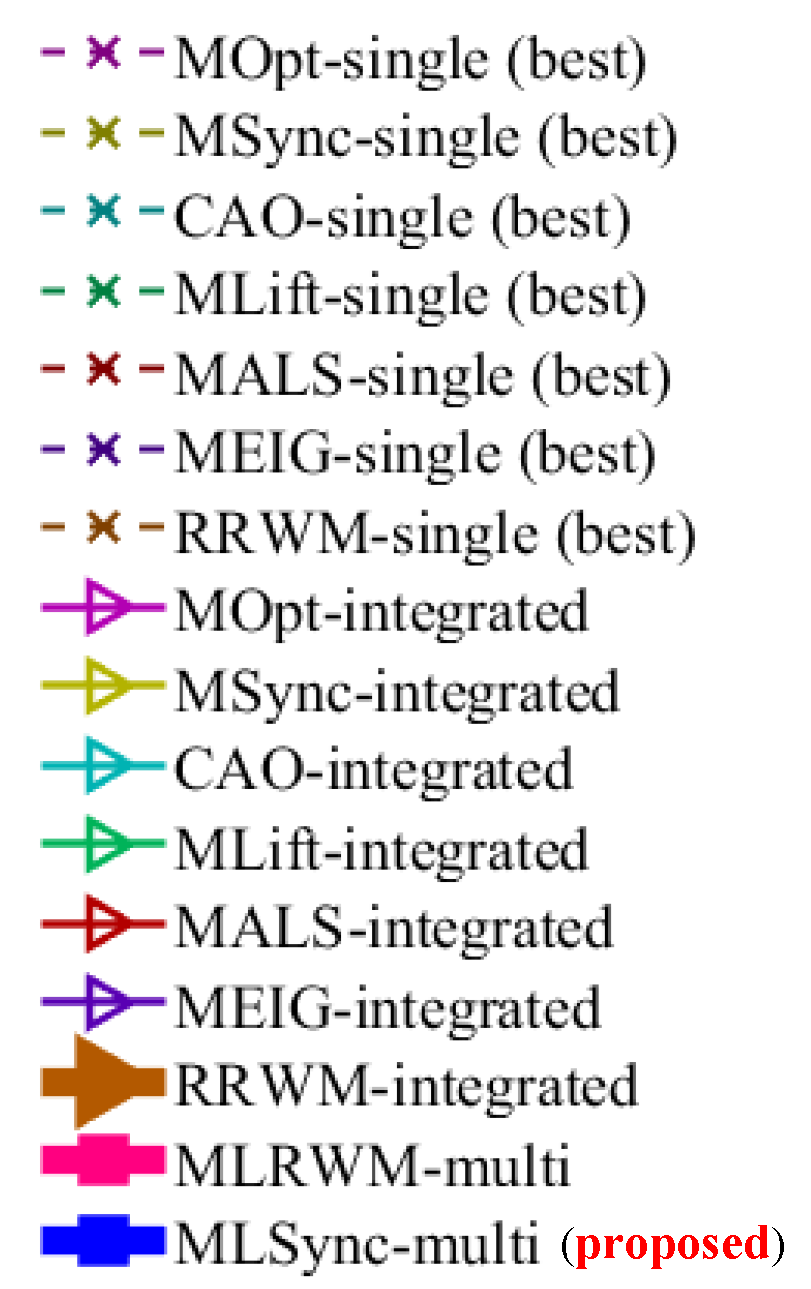}
	\caption{ Deformation experiments}   
  \end{subfigure}

  \begin{subfigure}[t]{\textwidth}
    \includegraphics[width=0.42\linewidth]{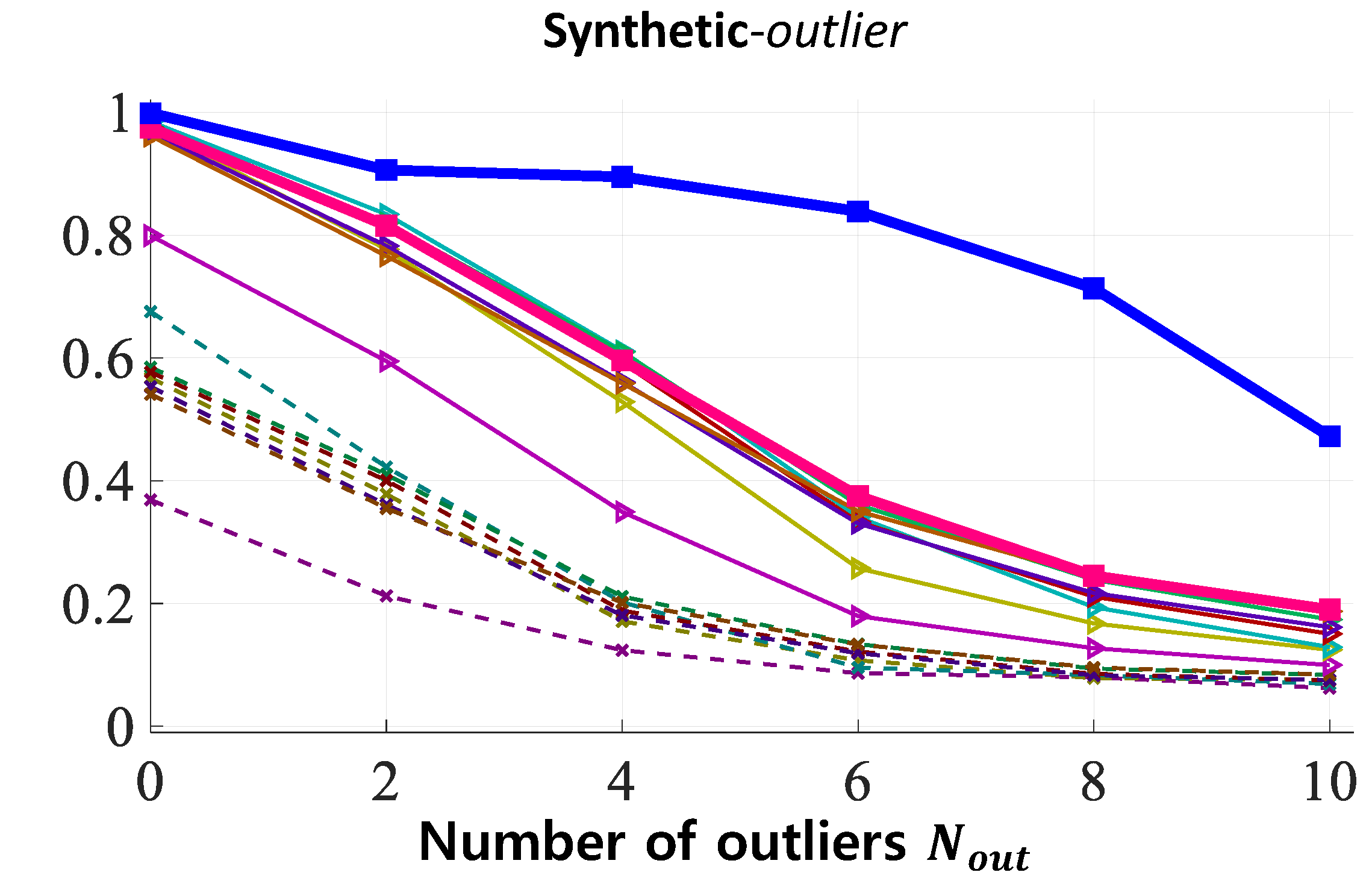}
    \includegraphics[width=0.42\linewidth]{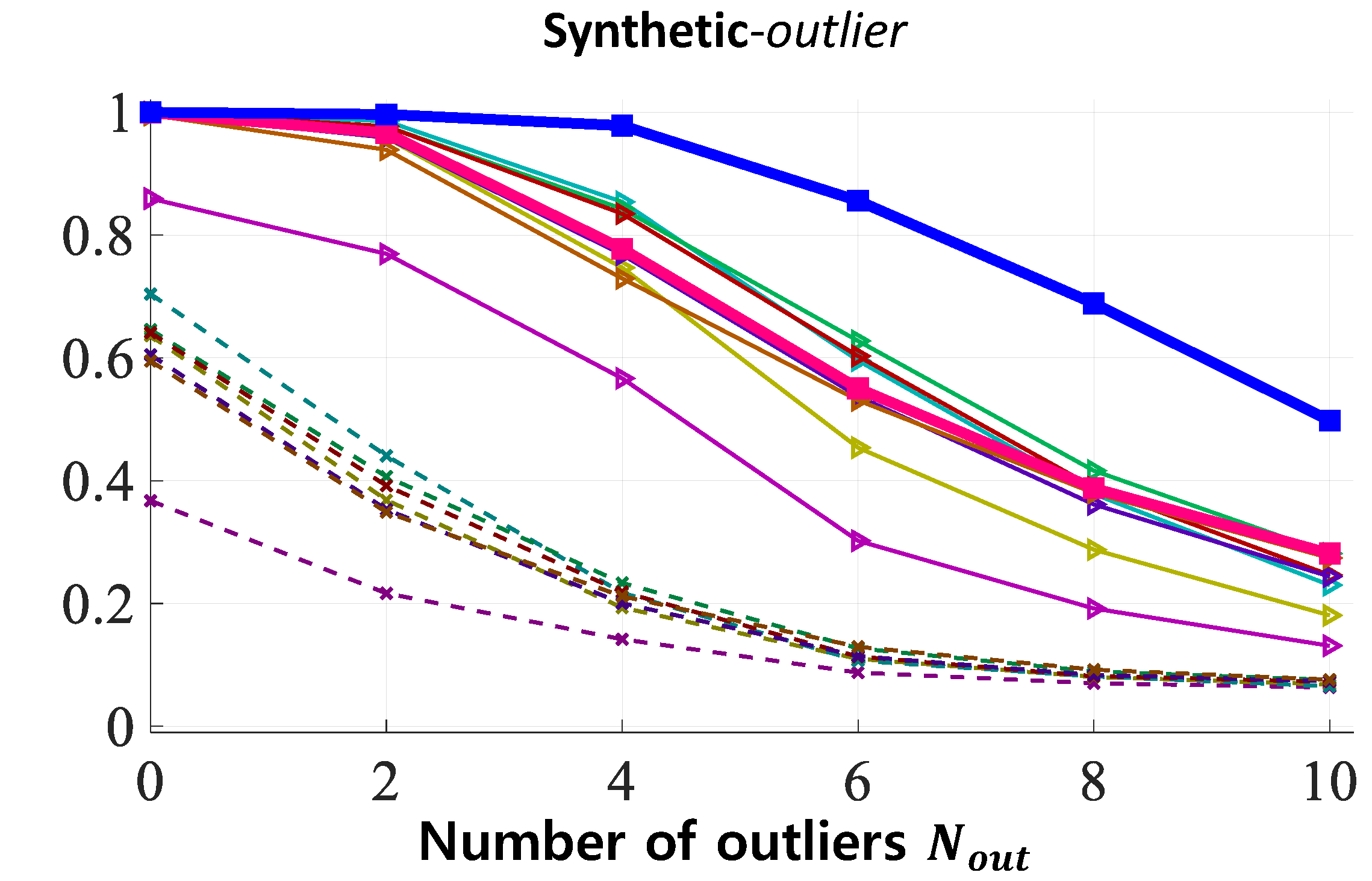}
    \includegraphics[width=0.15\linewidth]{./Figures/Figure_legend_2.png}
    \caption{ Outlier experiments} 
  \end{subfigure}

  \begin{subfigure}[t]{\textwidth}
    \includegraphics[width=0.42\linewidth]{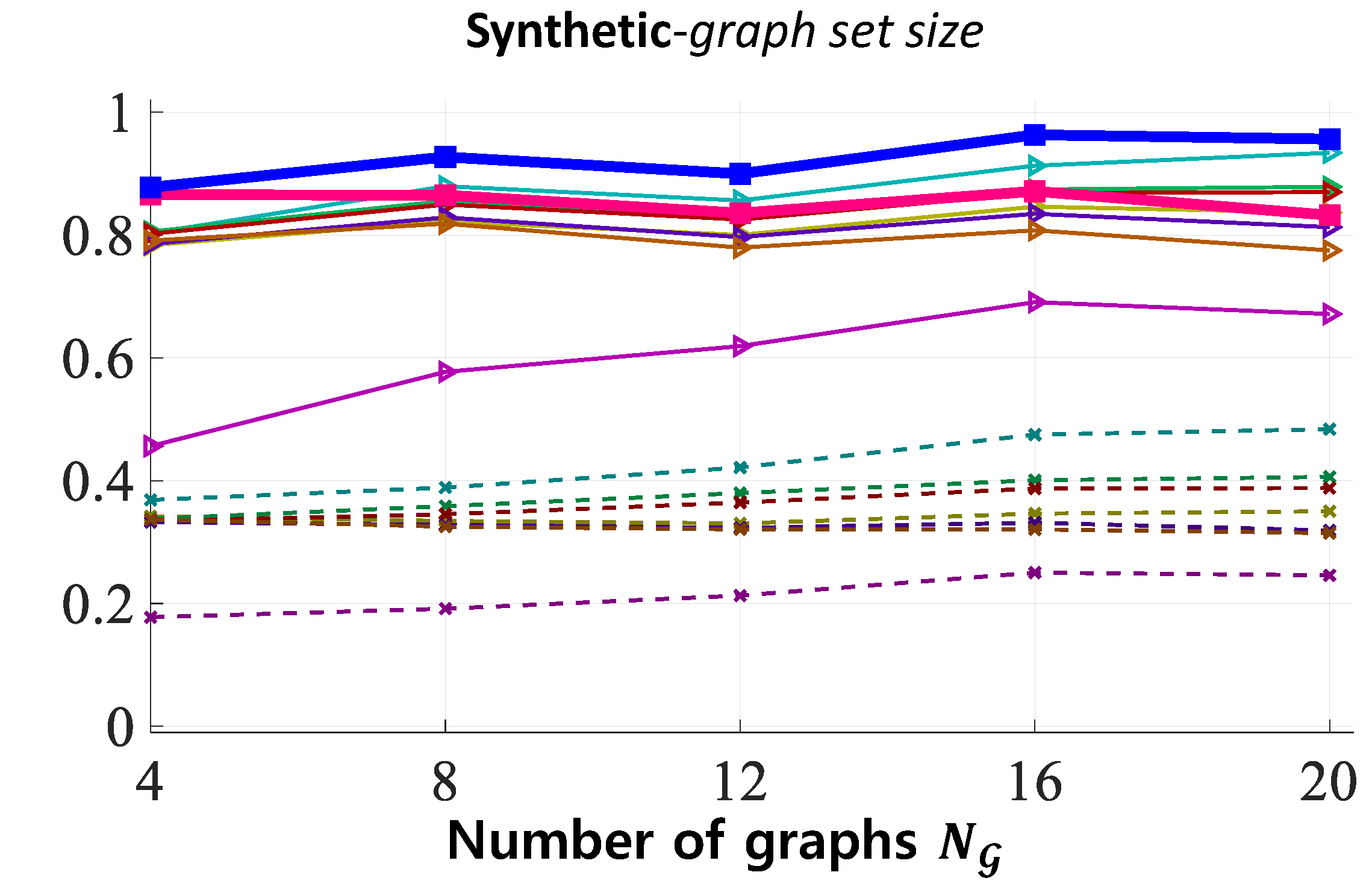}
    \includegraphics[width=0.42\linewidth]{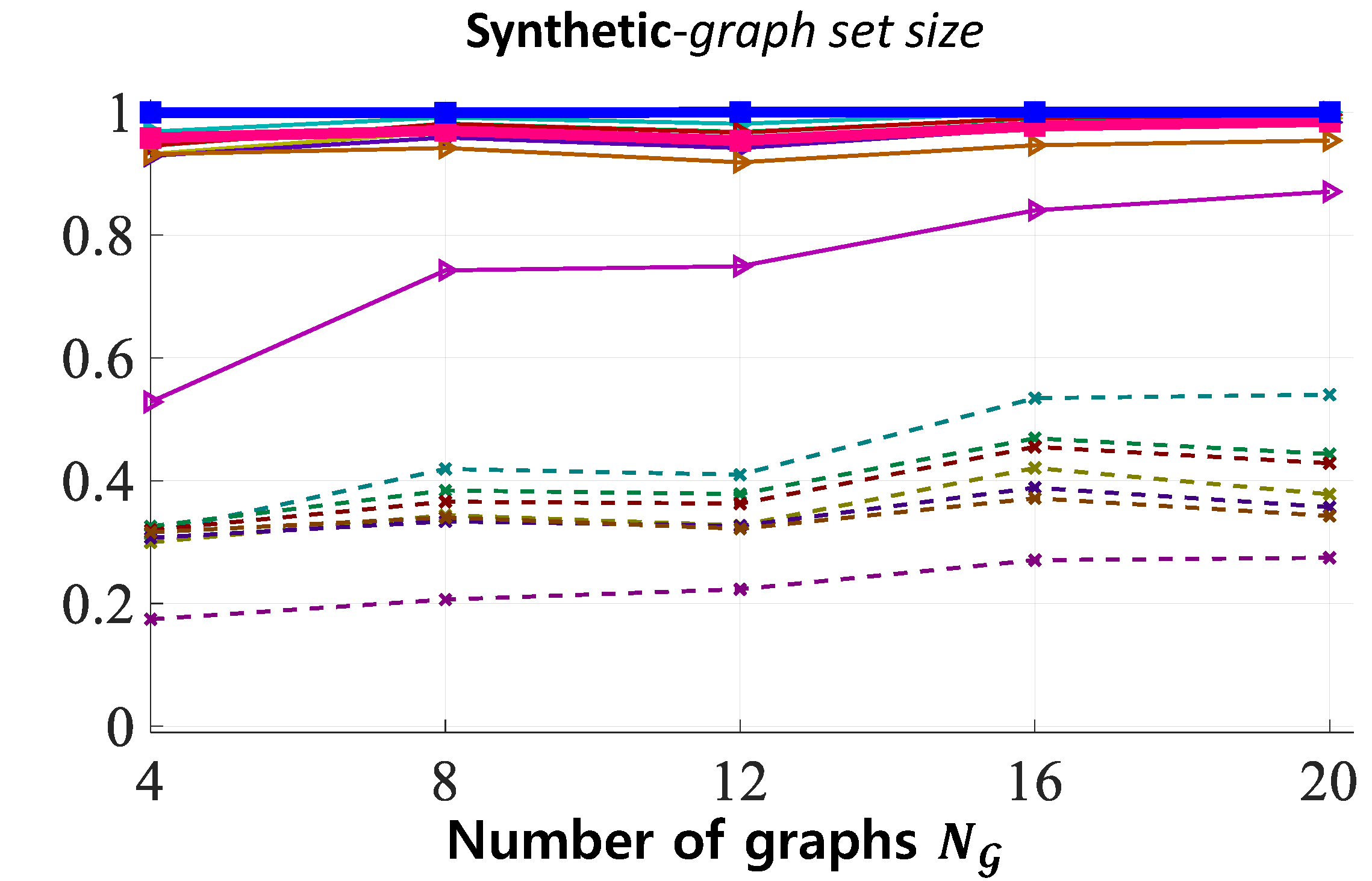}
    \includegraphics[width=0.15\linewidth]{./Figures/Figure_legend_2.png}
    \caption{ Graph set size experiments} 
  \end{subfigure}
	\caption[Synthetic graph matching results.]{ Synthetic graph matching results. Rows: each type of experiments -- deformation, outlier, and graph set size. Columns: number of attributes -- five and ten attributes. } 
	\label{fig_experiments_synthetic}
    \vspace{-10pt}
\end{figure*}

In this experiment, we evaluate our algorithm for synthetic graph matching problems.
To generate a set of synthetic graphs, we follow the experimental scheme that was represented in \cite{park2016multi,park2017multi}.
First, we construct a reference graph $\mathcal{G}_{0}$ that has several types of attributes $\mathcal{A}_{0}$.
Each attribute in $\mathcal{A}_{0}$ is defined by assigning randomly generated values with different variance values which reflect its distinctive information.
Then, we construct each graph $\mathcal{G}_{m}$ by adding randomly defined outlier vertices ${v}_{out}$ and attribute deformation that follows a Gaussian distribution $\mathcal{N}\left(0,{\epsilon}_{2}\right)$.
Finally, each affinity value in a intra-layer affinity matrix ${\left[\mathbf{P}_{lm}^{\alpha;\alpha}\right]}_{ia,jb}$ of a graph pair $\left(\mathcal{G}_{l},\mathcal{G}_{m}\right)$ is defined as follows:
\begin{equation}
\left[\mathbf{P}_{lm}^{\alpha;\alpha}\right]_{ia,jb} = \text{exp}(-{|(1-\beta) + \beta({a}^{\alpha}_{(l)ij}-{a}^{\alpha}_{(m)ab})|}_{2}^{2}/{\sigma}^{2}),
\end{equation} 
where ${a}^{\alpha}_{(l)ij}$ and ${a}^{\alpha}_{(m)ab}$ are randomly assigned attribute values, and $\sigma$ is a scaling factor.
$\beta$ is a control parameter that determines the variance of attributes, and is randomly selected in the interval $[0.1,1]$.
To utilize multiple attributes for single-layer graph matching algorithms, we first normalize the affinity matrices individually, and then aggregate the matrices.
By normalizing the affinity matrices before integration, it moderates the negative effects caused by different scales of multiple attributes.

Then, we performed three experiments: \textit{deformation}, \textit{outlier}, and \textit{graph set size}.
In the first experiment, each graph is generated with different magnitudes of deformation $\epsilon$, but other parameters are fixed.
On the other hand, in the outlier experiment, only the number of outliers ${N}_{out}$ is changed.
In the graph set size experiment, the number of graphs is increased while other parameters are fixed.
The details about parameter settings are presented in Table~\ref{tab_param_synthetic}.

Figure~\ref{fig_experiments_synthetic} represents the experimental results, where the postfix `-single', `-integrated', and `-multi' represent the results obtained using a single attribute, an integrated multiple attributes, and multiple attributes, respectively. 
Note that we only present the best result among all the `-single' results for each method.
As represented in Fig.~\ref{fig_experiments_synthetic}, the proposed method (`MLSync-multi') outperforms the pairwise algorithms based on the random walks framework, RRWM and MLRWM, in all experiments.
In particular, our algorithm shows improved performance than MLRWM when a set of graphs becomes larger, which reflects that the proposed concept of synchronization works well as we intended.
Moreover, the proposed algorithm exhibits comparable performance to the state-of-the-art multiple graph matching algorithms even in very noisy environments.


\begin{table*}[t!]
	\centering
	\caption{Attribute description}
	\label{tab_attribute_list}
	{
    	\small
		\begin{tabular}{l|l|l}
			\noalign{\hrule height 0.5pt} \hline
			Type of attributes			&		Abbreviation			& Description			     				\\ \hline	\hline
			\multirow{6}{*}{Appearance}	&		\textbf{USiD}			& \textit{Unary} SIFT descriptor difference   		\\ 
			&		\textbf{PSiD} 		& \textit{Pairwise} SIFT descriptor difference 				\\
			&		\textbf{CSiD}       & \textit{Concatenated} SIFT descriptor difference  	\\ 
			&		\textbf{UCoD}		& \textit{Unary} color histogram difference   		\\ 			
			&		\textbf{PCoD}		& \textit{Pairwise} color histogram difference 				\\
			&		\textbf{CCoD}		& \textit{Concatenated} color histogram difference	\\  \hline
			\multirow{3}{*}{Geometric}	&		\textbf{RDHD}			& Relative \textit{distance} histogram difference	\\ 
			&		\textbf{RAHD}		& Relative \textit{angle} histogram difference		\\ 
			&		\textbf{HARG}		& Relative \textit{distance} and \textit{angle} histogram difference (HARG~\cite{cho2013learning})   \\ \hline
			Multiple &		\textbf{Multi}	& Combination of all attributes   			\\ 
			\noalign{\hrule height 0.5pt} \hline
		\end{tabular}
	}
    \vspace{-12pt}
\end{table*}

\subsection{Performance evaluation on WILLOW dataset}
\label{image_experiments}


To define multi-attributed multiple graph matching problems, we use nine attributes as represented in Table~\ref{tab_attribute_list}.
The attributes can be roughly classified into two types: appearnce and geometric attributes.
The appearance attributes are defined using a SIFT descriptor~\cite{lowe2004distinctive} or an RGB color histogram. 
Each prefix represents the style of description.
For instance, the '\textit{Pairwise}' SIFT descriptor (PSiD) implies that each edge attribute is obtained by calculating the difference between the SIFT descriptors of two interest points.
Similarly, the '\textit{Concatenated}' color histogram difference (CCoD) means that each attribute is represented by concatenating two SIFT descriptors.
The last type of attributes, such as '\textit{Unary}' SIFT descriptor difference (USiD), means that each vertex attribute is defined by using a descriptor of each interest point; for this reason, the affinity matrix of this type has only unary affinity values (a diagonal matrix).
On the other hand, the geometric attributes are obtained from the HARG~\cite{cho2013learning}.
The HARG describes each pair of points by using relative \textit{distance} and \textit{angle} histograms.
Thus, we define the attributes, relative \textit{distance} histogram difference (RDHD) and relative \textit{angle} histogram difference (RAHD) by decomposing the HARG histogram.
Then, similar with the synthetic graph matching experiments in Sec.~\ref{synthetic_experiments}, we describe each affinity value in a intra-layer affinity matrix ${\left[\mathbf{P}_{lm}^{\alpha;\alpha}\right]}_{ia,jb}$ of a graph pair $\left(\mathcal{G}_{l},\mathcal{G}_{m}\right)$ is defined as follows: 
\begin{equation}
\left[\mathbf{P}_{lm}^{\alpha;\alpha}\right]_{ia,jb} = \text{exp}(-{|{a}^{\alpha}_{(l)ij}-{a}^{\alpha}_{(m)ab}|}_{2}^{2}/{\sigma}^{2}),
\end{equation} 
where ${a}^{\alpha}_{(l)ij}$ and ${a}^{\alpha}_{(m)ab}$ are attribute values obtained by following the definitions in Table~\ref{tab_attribute_list}, and $\sigma$ is a scaling factor.
To construct an integrated attributes, we also normalized the affinity matrices individually, and then aggregated the matrices.

To evaluate the proposed algorithm, we performed two experiments: \textit{outlier} and \textit{graph set size}.
The outlier experiments change only the number of outliers ${N}_{out}$ as a control parameter because the attribute deformation is not controllable in the real image dataset as opposed to the synthetic graph matching experiments.
Thus we use the outlier experiment for evaluating the robustness of the proposed algorithm in very noisy environments with deformation and outliers together.
In the graph set size experiments, the number of graphs is varied while other parameters are fixed.
The details about parameter settings are shown in Table~\ref{tab_param_WILLOW}.


\begin{table*}[t]
	\centering
	\vspace{-5pt}
    \caption{Parameter setting for the real image graph matching experiments}
	\label{tab_param_WILLOW}
	{
    	\small
		\begin{tabular}{l|c|c}
			\noalign{\hrule height 0.5pt} \hline			
			Experiments      & Varied parameter                    & Fixed parameters                                    \\ \hline		\hline
			Outlier          &    ${N}_{out}=0$ -- $10$     &  ${N}_{\mathcal{G}}=10$, ${N}_{att}=9$, ${N}_{in}=10$, ${\sigma}^{2}=0.3$  \\
			Graph set size	 &    ${N}_{\mathcal{G}}=4$ -- $20$    &  ${N}_{att}=9$, ${N}_{in}=10$, ${N}_{out}=2$, ${\sigma}^{2}=0.3$     \\ 
			\noalign{\hrule height 0.5pt} \hline	
		\end{tabular}
	}
    \vspace{-10pt}
\end{table*}

\begin{table*}[t!]
    \centering
    \caption{Performance evaluation on WILLOW dataset~(Varying the number of outliers). \textcolor{red}{\textbf{Red}} and \textcolor{blue}{\textbf{blue}} bold numbers indicate the best and the second-best performances.}
    \label{tab_experiments_WILLOW_outlier}
    \footnotesize 
    \begin{tabular}{m{1.3cm} m{2.8cm} l l l l l l}
          \noalign{\hrule height 0.5pt}\hline
          \multicolumn{2}{c}{\multirow{2}{*}{Methods}} & \multicolumn{6}{c}{Category}  \\ \cline{3-8} 
          &  & \multicolumn{1}{m{1.2cm}}{\textit{face}} & \multicolumn{1}{m{1.1cm}}{\textit{moto.}} & \multicolumn{1}{m{1.1cm}}{\textit{car}} & \multicolumn{1}{m{1.1cm}}{\textit{duck}} & \multicolumn{1}{m{1.1cm}}{\textit{wine.}} & \multicolumn{1}{m{1.1cm}}{\textbf{Avg.}} \\ \noalign{\hrule height 0.5pt}\hline
          \multirow{2}{*}{Pairwise} & RRWM~\cite{cho2010reweighted} 				& 62.49 & 50.66 & 47.59 & 42.27 & 52.51 & 51.11 \\ 
          & MLRWM~\cite{park2016multi,park2017multi}    & 63.55 & \color{blue}\textbf{52.71} & \color{blue}\textbf{50.32} & \color{blue}\textbf{43.69} & \color{blue}\textbf{54.49} & 52.95 \\ \hline
          \multirow{6}{*}{Multiple} & MOpt~\cite{yan2013joint,yan2015consistency} & 30.76 & 26.39 & 24.31 & 22.50 & 26.98 & 26.19 \\ 
          & MSync~\cite{pachauri2013solving} & 60.65 & 45.84 & 43.70 & 38.47 & 48.55 & 47.44 \\ 
          & CAO~\cite{yan2016multi,yan2014graduated} & 62.51 & 46.41 & 44.79 & 39.85 & 49.05 & 48.52 \\ 
          & MLift~\cite{chen2014near} & \color{red}\textbf{65.93} & 51.99 & 49.85 & 43.22 & 54.05 & \color{blue}\textbf{53.01} \\ 
          & MALS~\cite{Zhou_2015_ICCV} & 64.69 & 50.79 & 48.33 & 41.90 & 52.95 & 51.74 \\ 
          & MEIG~\cite{Maset_2017_ICCV} & 62.48 & 49.82 & 46.72 & 41.04 & 51.55 & 50.32 \\ \cline{2-8} 
          & MLSync (Proposed) & \color{blue}\textbf{64.95} & \color{red}\textbf{54.02} & \color{red}\textbf{51.59} & \color{red}\textbf{44.53} & \color{red}\textbf{55.06} & \color{red}\textbf{54.03} \\ \noalign{\hrule height 0.5pt}\hline
    \end{tabular}
    \vspace{-10pt}
\end{table*}



\begin{figure*}[t!]
  \begin{subfigure}[t]{\textwidth}
    \includegraphics[width=0.47\linewidth]{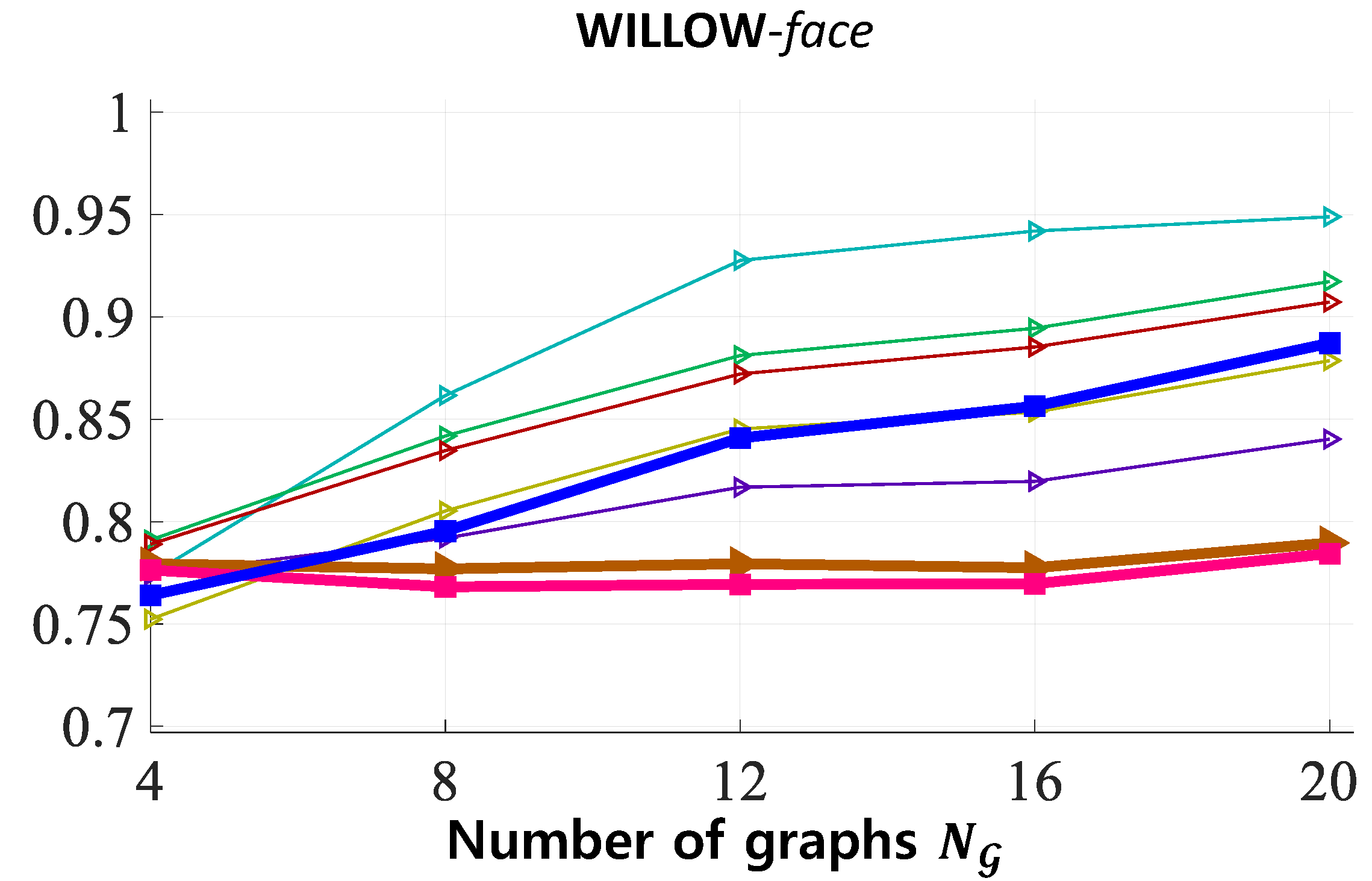}
    \includegraphics[width=0.47\linewidth]{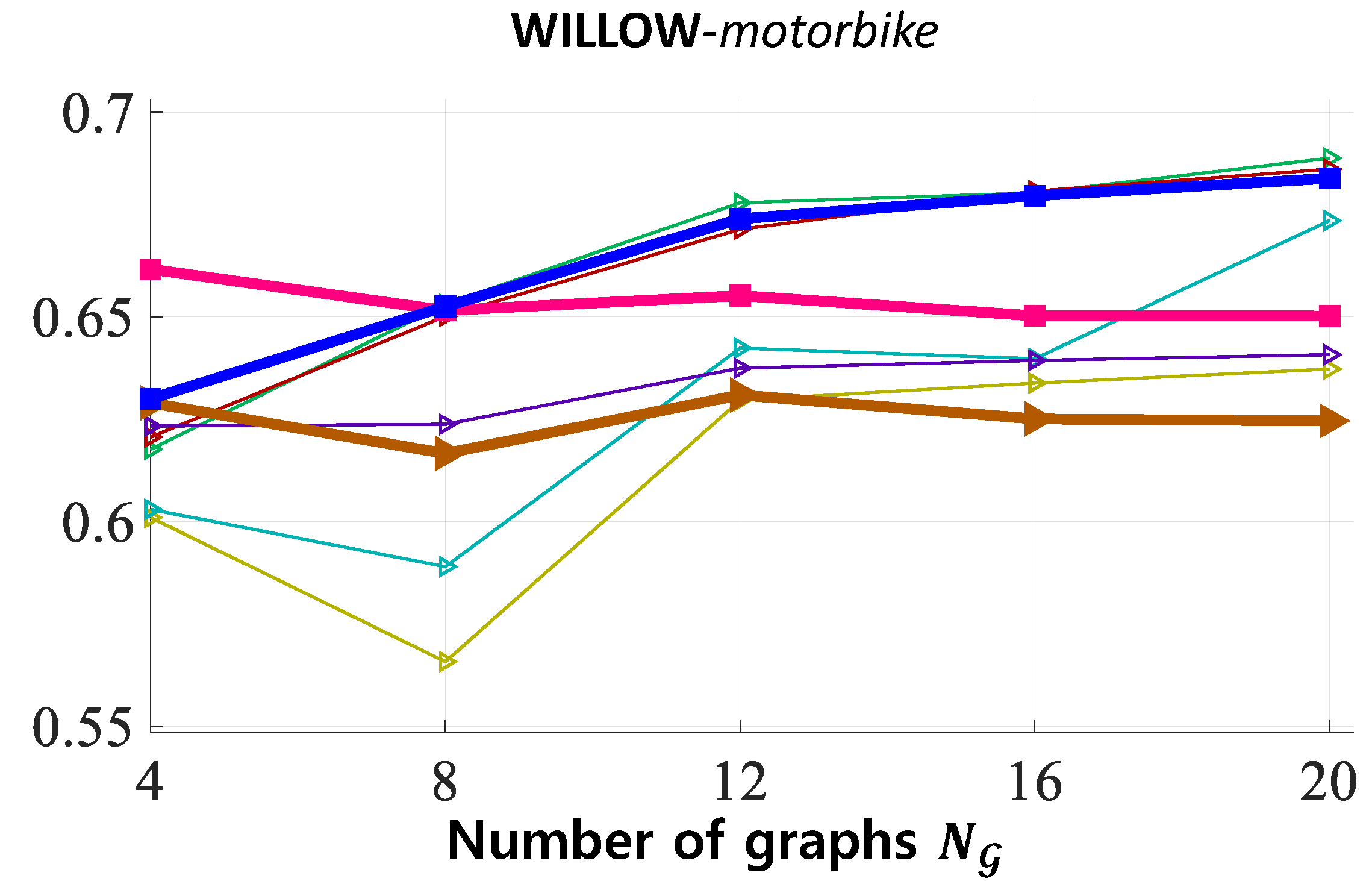}
  \end{subfigure}

  \begin{subfigure}[t]{\textwidth}
    \includegraphics[width=0.47\linewidth]{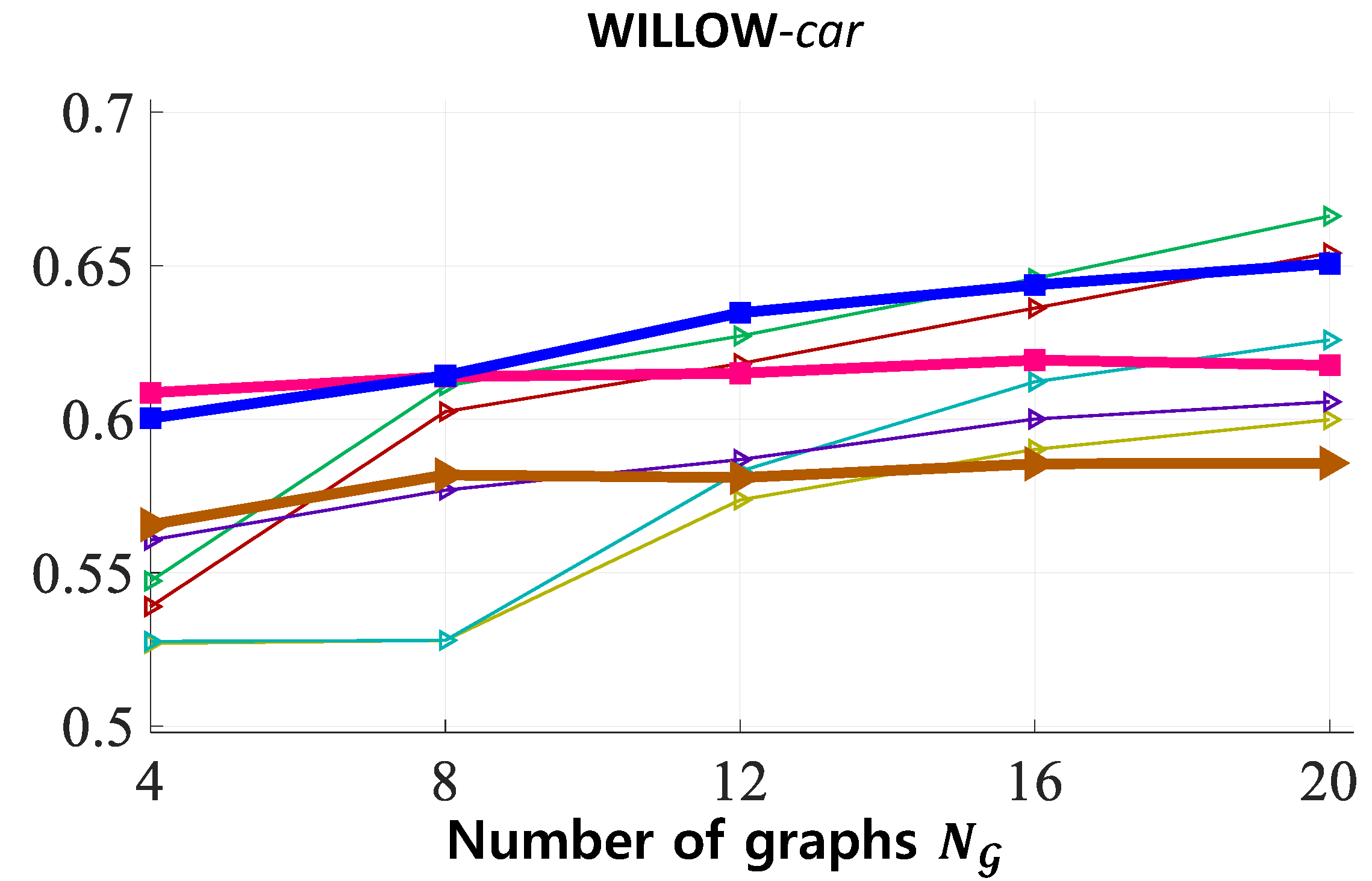}
    \includegraphics[width=0.47\linewidth]{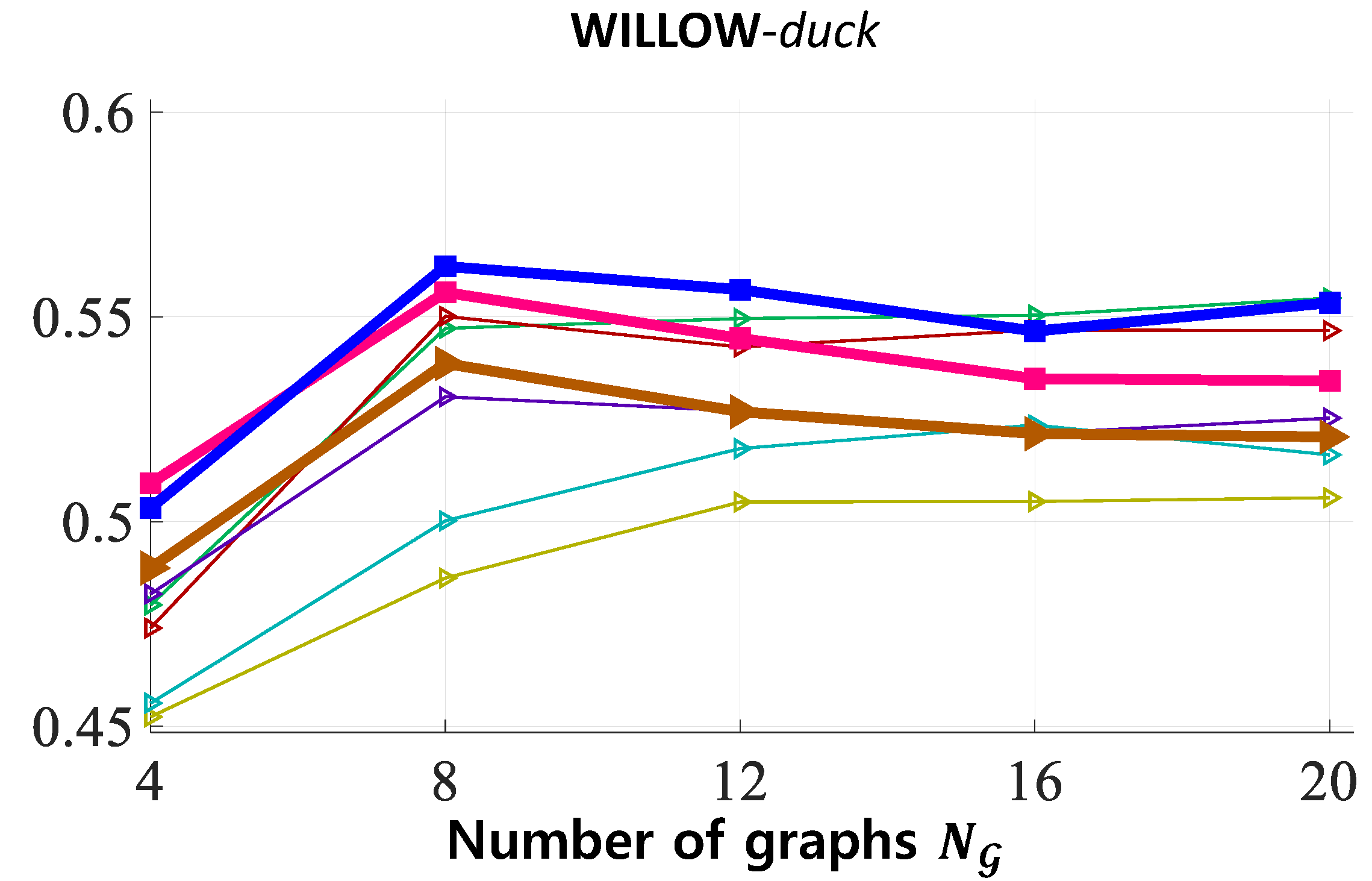}
  \end{subfigure}

  \begin{subfigure}[t]{\textwidth}
    \includegraphics[width=0.47\linewidth]{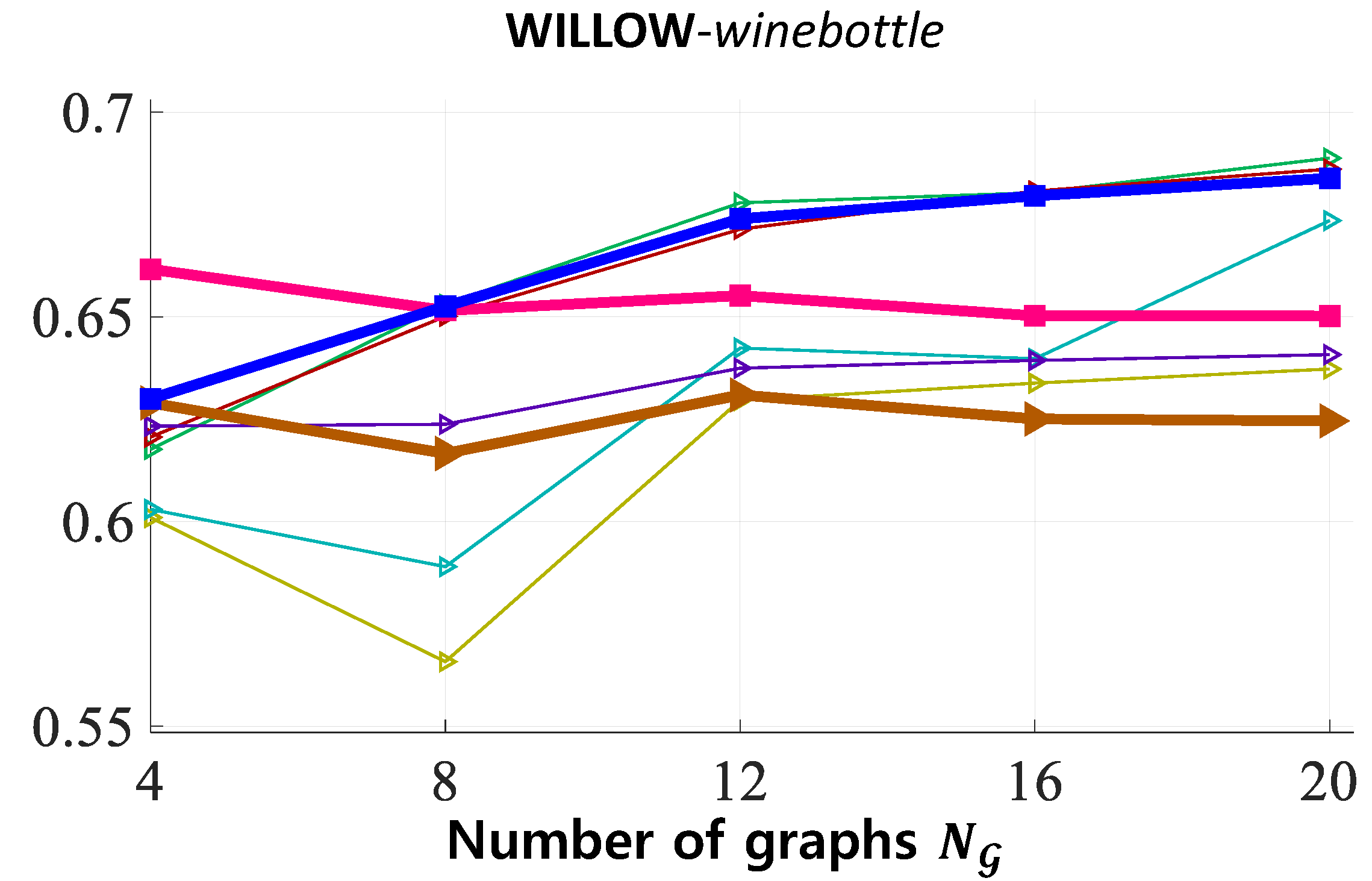}
    \includegraphics[width=0.44\linewidth]{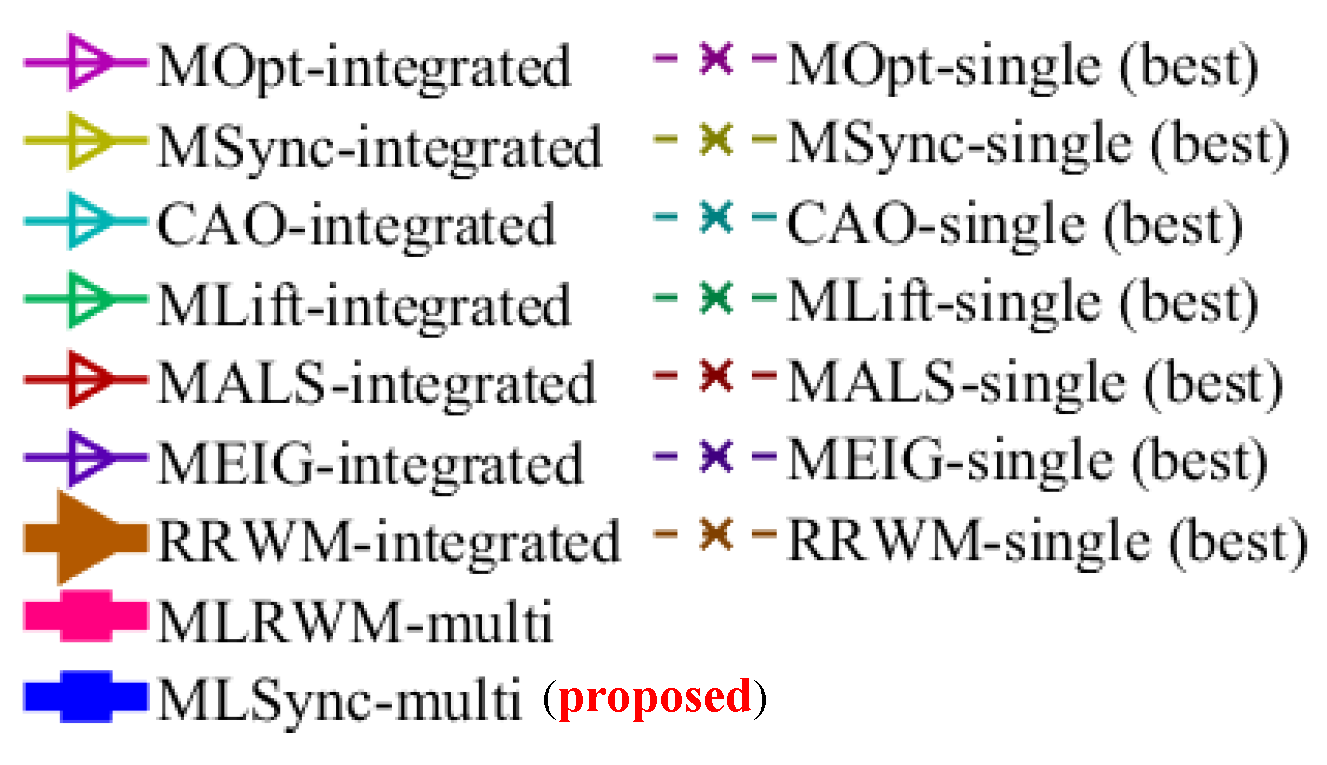}
  \end{subfigure}
	\caption{Performance evaluation on WILLOW dataset~(Varying the number of graphs)} 
	\label{fig_experiments_WILLOW_graph_set}
    \vspace{-10pt}
\end{figure*}



As represented in Tab.~\ref{tab_experiments_WILLOW_outlier} and Fig.~\ref{fig_experiments_WILLOW_graph_set}, the proposed algorithm also outperforms RRWM and MLRWM for all categories, as in the case of the synthetic graph matching experiments.
However, our algorithm represents relatively low accuracy in the \textit{face} category, but it is still comparable to other algorithms.
It is because the layer confidence measure is likely to fail to estimate correct confidence values more frequently than other categories, even though the proposed algorithm synchronizes the confidence values. 
We expect that the confidence measure could be improved by using more sophisticated methods such as machine learning methods.
We remains this issue as a future work.



\section{Conclusion}
\label{conclusion}

In this paper, we proposed a novel multi-attributed multiple graph matching algorithm based on the multi-layer random walks synchronization.
The algorithm aims to solve multiple graph matching problems in complicated environments by using multiple attributes that are represented in a set of multi-layer structures.
To improve the matching consistency among graphs, we proposed a random walks synchronization process which leads random walkers to consistent matching candidates.
In our extensive experiments, the proposed algorithm exhibits very robust, consistent, and accurate performances over the state-of-the-art algorithms.


\bibliographystyle{splncs}
\bibliography{egbib}
\end{document}